%% file: main.tex
\PassOptionsToPackage{table}{xcolor}
\documentclass[10pt]{article}
\usepackage[preprint]{tmlr}

\usepackage[utf8]{inputenc}
\usepackage[T1]{fontenc}
\usepackage{url}
\usepackage{nicefrac}
\usepackage{microtype}
\usepackage{xspace}

\usepackage{amsmath,amssymb,mathtools}

\usepackage{graphicx}
\usepackage{booktabs}
\usepackage{tabularx}
\usepackage{multirow}
\usepackage{makecell}
\usepackage{adjustbox}
\usepackage{caption}
\usepackage{subcaption}
\usepackage{longtable}
\usepackage{tikz}
\usetikzlibrary{arrows.meta,positioning,fit}
\usepackage{xcolor}

\usepackage{algorithm}
\usepackage{algorithmic}

\usepackage{siunitx}
\usepackage{float}     % enables [H]
\usepackage{placeins}  % \FloatBarrier
\usepackage{flafter}  % Place floats after their source position.
\usepackage{needspace}
\usepackage[hypertexnames=false]{hyperref}

\newcommand{\cmark}{\ensuremath{\checkmark}}
\newcommand{\xmark}{\ensuremath{\times}}

\definecolor{passblue}{HTML}{DCEEFF}
\definecolor{failred}{HTML}{FFE5E5}
\definecolor{failgray}{HTML}{F0F0F0}

\newcommand{\method}{\textsc{GUIDE}\xspace}
\newcommand{\Q}{\mathsf{Q}}
\newcommand{\U}{\mathsf{U}}
\newcommand{\Ind}{\mathsf{I}}
\newcommand{\Simi}{\mathsf{S}}
\newcommand{\Emin}{\mathsf{E}_{\min}}

\newcommand{\eps}{\varepsilon}

\newcommand{\Dtest}{D_{\text{test}}}
\newcommand{\DtrainOne}{D_{\text{train1}}}
\newcommand{\DtrainTwo}{D_{\text{train2}}}
\newcommand{\Dgen}{D_{\text{generated}}}

\title{Beyond Realism: Evaluating Generative Data Through Downstream Tasks and Distributional Diagnostics}

\author{%
  \name Cosmina Simionescu \email cs843@exeter.ac.uk \\
  \addr Department of Computer Science\\
  University of Exeter\\
  United Kingdom
  \AND
  \name Fabrizio Costa \email f.costa@exeter.ac.uk \\
  \addr Department of Computer Science\\
  University of Exeter\\
  United Kingdom
}
\def\month{09}
\def\year{2026}

\begin{document}
\raggedbottom

\maketitle
\let\AND\undefined
\begin{abstract}
Evaluating generative data is difficult because resemblance to real data does not guarantee downstream value, different metrics can favor different generators, and finite samples make those comparisons uncertain. We propose \textbf{\method} (\emph{Generative Utility and Indistinguishability Diagnostic Evaluation}), a protocol for evaluating generated datasets through downstream tasks. GUIDE assesses whether generated samples (1) preserve predictive signal, (2) improve performance beyond an equally sized addition of real data, (3) can be distinguished from real samples, and (4) mix locally with real samples within task classes. This diagnostic view distinguishes predictive redundancy from distributional resemblance. We derive conditional finite-sample bounds to motivate metric-specific acceptance screens and compare generators while accounting for evaluation uncertainty. In controlled perturbation experiments, we test GUIDE against pre-specified failure hypotheses. As the share of training-set copies increases, local similarity rises while augmentation gains decline. This divergence shows that greater resemblance to real data does not necessarily mean greater value for downstream learning. We also demonstrate full or partial applications of GUIDE to graph, image, and text generators.
\end{abstract}

\section{Introduction}
\label{sec:intro}

How should we rank generative models when different evaluation metrics
capture different properties of their samples, and when the estimates
of these properties are subject to uncertainty? A generator may produce
realistic samples while covering only a small part of the real
distribution, whereas another may achieve broader coverage at the
expense of sample fidelity. Different evaluation criteria can therefore
favor different generators, and their relevance depends on the intended
application~\citep{theis2016note,sajjadi2018assessing}.
A further consideration is how strongly the available evaluation data
support the observed differences between generators.

Existing metrics address complementary aspects of generative model
evaluation. Fr\'echet Inception Distance
(FID)~\citep{heusel2017gans} and kernel-based measures such as Kernel
Inception Distance (KID) and maximum mean discrepancy
(MMD)~\citep{binkowski2018demystifying,gretton2012kernel}
quantify discrepancies between real and generated distributions in a
chosen feature representation. Precision--recall
metrics~\citep{sajjadi2018assessing,kynkaanniemi2019improved,simon2019revisiting}
and density--coverage~\citep{naeem2020reliable} distinguish aspects of
sample fidelity and distributional coverage. Although these measures
provide valuable information about the relationship between real and
generated distributions, distributional resemblance does not necessarily
imply usefulness for downstream learning. Training-set copying
illustrates this distinction: generated samples may closely resemble
real data without extending beyond the examples already available to
the learner~\citep{meehan2020non,alaa2022faithful,jiralerspong2023feature}.

Prediction-based evaluation offers a complementary perspective by
examining the information that generated samples provide for a
downstream task. Training a predictor on generated samples and
evaluating it on held-out real data tests whether task-relevant
structure transfers between the two
distributions~\citep{shmelkov2018good,hyland2017real,yoon2019timegan}.
\citet{shmelkov2018good} also examine the predictive benefit of
augmenting real training data with generated samples. Classifier
two-sample tests address a different question by measuring how readily
samples from two distributions can be
distinguished~\citep{lopezpaz2016revisiting}.
These comparisons capture different properties of generated data:
samples that support accurate prediction when used alone need not
provide additional predictive value when combined with existing
real training data.

Beyond their diagnostic role, learning-based evaluation metrics
offer an opportunity to draw on established results from statistical
learning theory. When an evaluation metric is formulated in terms
of predictive performance or discrimination between distributions,
we can use appropriate learning-theoretic bounds to characterize
the uncertainty associated with its estimation. This provides a basis
for distinguishing a favorable empirical score from one for which
the available data provide sufficient support at a prescribed level
of performance. Incorporating this distinction into model comparison
allows us to consider both the observed properties of generated data
and the uncertainty associated with their evaluation.

\begin{figure}[!htbp]
\centering
\includegraphics[width=0.9 \linewidth]{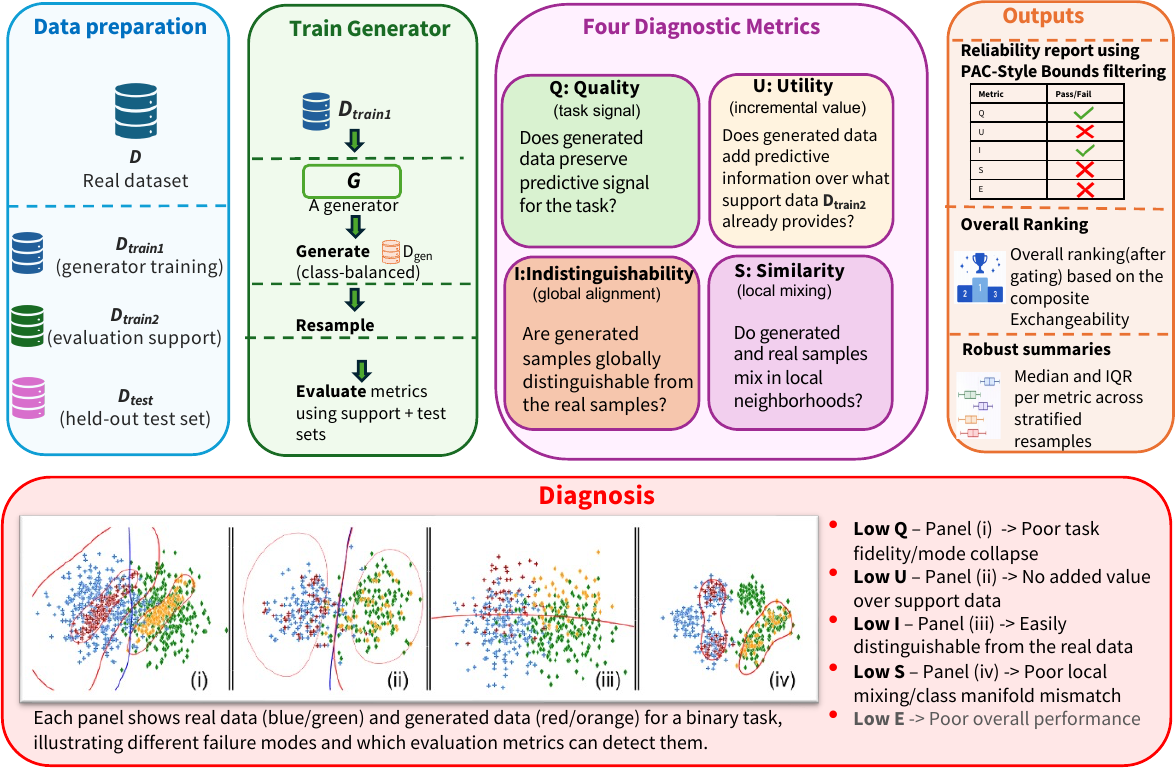}
\caption{\textbf{\method pipeline.}
Real data are partitioned into generator-training, additional
real-data, and held-out evaluation sets. Generated samples are
evaluated through four complementary diagnostic metrics.
Conditional uncertainty bounds motivate metric-specific acceptance
screens. Generators are grouped by the number of failed screens and
ranked within each group through Monte Carlo comparisons of their
four metric scores.
The bottom panels illustrate different discrepancies between real
and generated data that motivate the complementary diagnostics.}
\label{fig:guide-pipeline}
\end{figure}

We propose \textbf{\method} (\emph{Generative Utility and
Indistinguishability Diagnostic Evaluation}), a task-grounded
protocol for diagnosing and ranking generative models through
the samples they produce (Figure~\ref{fig:guide-pipeline}).
The protocol comprises four complementary metrics.
\textit{Quality} assesses whether generated samples preserve
the information needed to perform a selected prediction task.
\textit{Utility} measures the incremental predictive value of
generated samples relative to an equally sized addition of real
training data.
\textit{Indistinguishability} measures how difficult it is for a
classifier to distinguish real from generated samples.
\textit{Similarity} measures the extent to which real and generated
samples mix within class-conditional local neighborhoods.
Task labels may be supplied by the data or defined through an
explicit proxy task (Appendix~\ref{appendix:proxy-supervision}).

\FloatBarrier
The formulation of these metrics allows us to derive conditional
finite-sample lower bounds by combining domain adaptation
bounds~\citep{ben2010theory}, complexity-based generalization
bounds~\citep{bartlett2002rademacher,mohri2018foundations},
and concentration inequalities for sampling with independent
observations or without replacement
\citep{hoeffding1963probability,serfling1974}. For Quality and Utility, we relate
predictive risks across real and generated distributions.
For Indistinguishability, we account for uncertainty in the
discriminator's performance. For Similarity, we characterize
uncertainty in local proportions of generated samples under an explicit
neighborhood sampling model. Each metric is associated with an
acceptance threshold, and its lower bound determines whether
the available evidence is sufficient to support that threshold
under the stated assumptions.

These bounds motivate an uncertainty-aware approach to ranking
generative models. Rather than constructing a ranking solely from
empirical metric values, we first group generators according to
the number of metrics whose lower bounds satisfy their respective
acceptance thresholds. Within each group, we compare the four
metric profiles through Monte Carlo pairwise dominance, incorporating
variability across evaluation resamples
(Algorithm~\ref{alg:main-ranking}).
The ranking is accompanied by the individual metric scores and
screen outcomes, allowing us to identify which properties account
for differences between generators.

We examine the diagnostic behavior of \method{} through controlled
experiments involving copying, label corruption, feature noise,
and mode collapse, comparing the observed metric responses with
pre-specified failure hypotheses.
The results illustrate why complementary metrics are needed:
increasing the proportion of training-set copies reduces
augmentation gains more consistently than generated-only predictive
performance, even as local mixing between real and generated samples
increases. Finally, we demonstrate the application of the protocol
to molecular graph and image and text generators.
\section{Task-Grounded Evaluation Problem}
\label{sec:problem}

An evaluation instance $(G,D,T)$ consists of a generator $G$,
a labeled reference dataset $D=\{(x_j,y_j)\}$, and a classification
task $T$ with performance measure $s(\cdot)$, where higher values
indicate better predictive performance. The task specifies the
properties of the generated samples that we wish to evaluate.

We partition $D$ into three disjoint subsets: a primary training
set $\DtrainOne$, an additional real-data set $\DtrainTwo$, and
a held-out test set $\Dtest$. In learned-generator experiments,
we train or fine-tune $G$ on $\DtrainOne$ and draw a generated
dataset $\Dgen$ from it. The primary training set also serves
as the real-only reference for evaluating classifier performance.
The second training set provides additional real samples against
which we compare the predictive contribution of an equally sized
set of generated samples. The held-out test set is reserved for
evaluating the task classifiers after feature and model fitting.

In the controlled failure experiments, we initially draw $\Dgen$
from the same known task distribution as the real splits.
We subsequently perturb $\Dgen$ to examine how the evaluation
metrics respond to controlled changes such as copying, label
corruption, feature noise, and mode collapse.

Throughout, pooling datasets, denoted by $\cup$, refers to
concatenating their observations while retaining duplicates.
The resulting evaluation scores are interpreted with respect
to the selected task, feature representation, data partition,
and classifier configuration.

\section{\method Protocol}
\label{sec:method}

\method{} constructs four complementary diagnostic metrics,
three of which are derived from classifier performance scores
obtained under different training and evaluation configurations.
The fourth metric examines local mixing between real and
generated samples within task classes. This formulation allows
us to draw on established theoretical results to derive conditional
finite-sample bounds and incorporate uncertainty into the
comparison of generative models.

\paragraph{Classifier performance comparisons.}
Let $h_A$ denote a task classifier trained on dataset $A$.
We train classifiers under four training-data configurations
and evaluate them on the same held-out real test set $\Dtest$,
using the same performance measure $s$:
\[
\begin{aligned}
a_r &= s(h_{\DtrainOne}), &
a_g &= s(h_{\Dgen}), \\
a_{rr} &= s(h_{\DtrainOne\cup\DtrainTwo}), &
a_{rg} &= s(h_{\DtrainOne\cup\Dgen}).
\end{aligned}
\]

The score $a_r$ establishes the real-only reference performance,
while $a_g$ measures the predictive performance achieved when
training exclusively on generated samples. The scores $a_{rr}$
and $a_{rg}$ measure predictive performance after augmenting
the primary training set with additional real and generated
samples, respectively.

In addition to these four task-classifier performance scores,
we train a separate binary classifier to distinguish real
samples from $\DtrainOne$ and generated samples from $\Dgen$,
following the classifier two-sample-test
approach~\citep{lopezpaz2016revisiting}.
We denote its balanced accuracy on the discriminator evaluation
set by $a_d$.

The resulting five classifier performance scores,
\[
(a_r,a_g,a_{rr},a_{rg},a_d),
\]
are used to construct Quality, Utility, and Indistinguishability.
Similarity is computed separately from the distribution of
real and generated samples in local neighborhoods.

\paragraph{Resampling and classifier configuration.}
We repeat the evaluation procedure over multiple resamples,
keeping the classifier configuration and performance measure
fixed within each comparison. In each resample, we compute
the five classifier performance scores and construct the
class-conditional neighborhoods required for Similarity. Resampling indices are selected independently of task labels.
In the label-corruption experiments, we retain the same indices
and reconstruct the class-conditional neighborhoods using
the perturbed labels.

The task classifier serves as an evaluation instrument rather than
as the model being optimized. We do not impose a minimum absolute
predictive score; instead, we compare its performance across the
training-data configurations. The resulting metrics nevertheless
depend on the classifier's ability to learn the relevant task
structure.

The molecular and image benchmarks use ExtraTrees~\citep{geurts2006extratrees}
as the task classifier. The controlled experiments use $5$-nearest
neighbors. In both cases, classifier performance is measured using F1
on the held-out real test set.
Appendix~\ref{appendix:default-classifiers} provides the
classifier configurations and experiment-specific overrides.

\phantomsection\label{sec:utility}
\paragraph{Quality and Utility.}
Quality and Utility are derived from the four task-classifier
performance scores and examine complementary aspects of the
predictive value of generated data.

\textit{Quality} measures how well a classifier trained exclusively
on generated samples performs relative to a classifier trained
on real samples:
\[
\Q=\mathrm{clip}_{[0,1]}\!
\left(\frac{a_g}{a_r+\eps}\right),
\]
where $\eps>0$ is a numerical stabilizer.

A high Quality score indicates that generated samples preserve
task-relevant structure sufficiently well to support classification
on real data. However, strong generated-only performance does not
necessarily imply that these samples provide additional predictive
value when combined with existing real training data.

\textit{Utility} addresses this distinction by comparing the
predictive gains obtained from augmenting the primary training
set with generated and additional real samples. We define
\[
\Delta_{\mathrm{generated}}=a_{rg}-a_r,
\qquad
\Delta_{\mathrm{real}}=a_{rr}-a_r.
\]

Utility measures the generated-data gain relative to the
matched real-data gain:
\[
\U=\mathrm{clip}_{[0,1]}\!
\left(
\frac{\Delta_{\mathrm{generated}}}
     {\Delta_{\mathrm{real}}}
\right),
\qquad
\Delta_{\mathrm{real}}>\tau.
\]

We set $\tau=10^{-3}$ to avoid normalization by a negligible
or non-positive reference gain. When
$\Delta_{\mathrm{real}}\leq\tau$, normalized Utility is
unavailable, and we report $\Delta_{\mathrm{generated}}$
separately as Utility Gain.

Together, Quality and Utility distinguish between the ability
of generated samples to support classification independently
and their ability to improve classification when added to
existing real training data.

\paragraph{Indistinguishability.}
Quality and Utility assess the predictive properties of generated
samples with respect to the selected task. Indistinguishability
instead examines how readily a classifier can distinguish
real samples from generated samples.

Using the balanced accuracy $a_d$ of the discriminator,
we define
\[
\Ind=1-2|a_d-0.5|.
\]

The transformation maps chance-level classification performance
to one and perfect separation, including perfect separation
with reversed predictions, to zero. A high Indistinguishability score indicates that the selected
discriminator has difficulty separating real and generated
samples in the chosen feature representation. Its interpretation
therefore depends on both the representation and the capacity
of the discriminator.

\paragraph{Similarity.}
While Indistinguishability assesses how readily a classifier
can separate real and generated samples, Similarity examines
how these samples mix locally within each task class.

Let $\mathcal C$ denote the set of task classes. For each class
$c\in\mathcal C$, we pool the $N_c$ resampled observations in
$\DtrainOne^c\cup\Dgen^c$. For each observation $j$, let $p_{cj}$
denote the proportion of generated samples among its $k$ nearest
neighbors, excluding the observation itself, with
$1\leq k<N_c$.

We quantify local mixing using binary entropy:
\[
H_2(p)=-p\log_2p-(1-p)\log_2(1-p),
\]
with $H_2(0)=H_2(1)=0$.

The class-specific Similarity score is obtained by averaging
the neighborhood entropies:
\[
S_c=\frac{1}{N_c}\sum_{j=1}^{N_c}H_2(p_{cj}).
\]

We aggregate the class-specific scores using their geometric mean:
\[
\Simi=\operatorname{gmean}_{c\in\mathcal C}S_c.
\]

The entropy is highest when real and generated samples are
equally represented in a neighborhood and lowest when a
neighborhood contains exclusively real or exclusively generated
samples. The geometric mean penalizes low Similarity in
individual task classes. High Similarity indicates local mixing but does not necessarily
imply that generated samples are novel, since copies of real
training samples may also increase local mixing. We additionally
report the proportion of generated samples in each class pool
to aid interpretation of the entropy scores.

Table~\ref{tab:classifier-metrics} summarizes the four diagnostic
metrics, the comparisons from which they are constructed, and
the interpretation of low scores.

\begin{table}[H]
\centering
\caption{The four diagnostic metrics of \method{}, their
underlying comparisons, and the interpretation of low scores.}
\label{tab:classifier-metrics}
\small
\begin{tabularx}{\linewidth}{@{}l X X@{}}
\toprule
\textbf{Metric} &
\textbf{Underlying comparison} &
\textbf{Interpretation of a low score} \\
\midrule

Quality &
Generated-only versus real-only classifier performance on
held-out real data. &
Generated samples support weaker predictive performance
than real samples. \\

\addlinespace[2pt]

Utility &
Predictive gain from adding generated samples relative to
the gain from adding an equally sized set of real samples. &
Generated samples provide limited incremental predictive
value relative to additional real data. \\

\addlinespace[2pt]

Indistinguishability &
Balanced accuracy of a classifier trained to distinguish
real from generated samples. &
Real and generated samples are readily distinguishable
in the chosen representation. \\

\addlinespace[2pt]

Similarity &
Binary entropy of the proportions of real and generated
samples in class-conditional nearest-neighbor neighborhoods. &
Real and generated samples exhibit limited local mixing
within task classes. \\

\bottomrule
\end{tabularx}
\end{table}

\paragraph{Uncertainty bounds and metric screening.}
The empirical values of the diagnostic metrics depend on
the available evaluation samples. An advantage of constructing
metrics from classifier performance scores is that established
theoretical results can be used to characterize uncertainty
in the quantities from which these metrics are derived.

For Quality and Utility, we use domain adaptation
bounds~\citep{ben2010theory} to relate classification risks
across real and generated distributions. Complexity-based
generalization bounds~\citep{bartlett2002rademacher,mohri2018foundations}
and Hoeffding's inequality~\citep{hoeffding1963probability}
provide controls on estimation error under the corresponding
capacity and sampling assumptions.

For Indistinguishability, we bound deviations between empirical
and population discriminator accuracy. For Similarity, we use
concentration inequalities for sampling without
replacement~\citep{serfling1974} to bound deviations between
observed and underlying proportions of generated samples
in local neighborhoods under an explicit sampling model.
These bounds are then propagated through the entropy and
aggregation operations.

For each metric $M\in\{\Q,\U,\Ind,\Simi\}$, let $\widehat M$
denote its empirical estimate and $M^\star$ its population
counterpart. Appendix~\ref{app:theory} derives conditional
finite-sample lower bounds $L_M$ satisfying
\[
\Pr\left(M^\star\geq L_M\right)\geq 1-\delta
\]
under the assumptions associated with each bound. Simultaneous
claims across metrics, generators, and resamples require an
appropriate allocation of failure probabilities.

Let $\theta_M$ denote a prescribed acceptance threshold
for metric $M$. We define its screening outcome as
\[
b_M=
\begin{cases}
1, & L_M\geq\theta_M,\\
0, & L_M<\theta_M.
\end{cases}
\]

A metric passes its screen when its lower bound meets the
prescribed threshold. This distinguishes a favorable empirical
score from one for which the available evaluation samples
provide sufficient support under the stated assumptions.
A failed screen may therefore reflect either a low empirical
score or substantial estimation uncertainty. It does not
necessarily imply that the population score is below
the threshold.

The formal Quality and Utility bounds use classification
accuracy and require appropriate bounds on domain discrepancy
and classification risks. The population Utility bound additionally
requires a lower limit for the real-data gain exceeding $\tau$.
The Similarity bound requires the conditional sampling model
specified in Appendix~\ref{app:theory}.

In the main experiments, the task classifiers are evaluated using
F1, and the uncertainty screens use approximate capacity inputs,
including $\sqrt{2/n}$ for the task-classifier complexity and
$d_{\mathrm{disc}}=50$ for the discriminator capacity.
The experimental screens compare
\[
\widehat M-r_M
\]
with $\theta_M$, where $r_M$ denotes the corresponding
diagnostic radius. These screens are motivated by the theoretical
bounds but are not formal confidence bounds for the reported F1-based
scores or for configurations in which the required assumptions
have not been established.

\phantomsection\label{sec:quantile-based-estimation}
\paragraph{Diagnostic profiles and uncertainty-aware ranking.}
Evaluating each generator over multiple resamples yields an empirical
distribution for each diagnostic metric, summarized by its median
and interquartile range (IQR). We use these summaries together with
the screening outcomes to compare generators.

Generators with complete metric profiles are first grouped by the
number of failed screens, giving preference to those satisfying
more acceptance criteria. Within each group, we use a Monte Carlo
procedure to account for variability in the metric values.
The empirical quartiles are converted into approximate means
and standard deviations (Appendix~\ref{app:quantile-to-moment}),
which parameterize the distributions used for sampling metric values.

At each Monte Carlo iteration, a generator dominates another if
it matches or exceeds its performance on all four metrics and
exceeds it on at least one. The accumulated pairwise dominance
counts determine the ordering within each group.

Algorithm~\ref{alg:main-ranking} summarizes the ranking procedure,
with its full specification given in Appendix~\ref{appendix:ranking}.
We report the ranking alongside the individual metric profiles
and screen outcomes. The ranking is conditional on the selected
thresholds and the Monte Carlo sampling model; the dominance counts
are not confidence probabilities for a population ordering. When
normalized Utility is unavailable, no four-metric rank is assigned.

\begin{algorithm}[!htbp]
\caption{Uncertainty-aware ranking of generative models}
\label{alg:main-ranking}
\begin{algorithmic}[1]
\REQUIRE Generators, resampled metric scores, and screening thresholds.
\STATE Compute the median, IQR, and screen status for each
       generator and metric.
\STATE Group generators with complete profiles by the number
       of failed screens, in ascending order.
\FOR{each group}
  \STATE Convert the empirical quartiles of each metric into
         a pseudo-mean and pseudo-standard deviation.
  \STATE Initialize pairwise dominance counts.
  \FOR{each Monte Carlo iteration}
    \STATE Independently sample metric values from the corresponding
           normal distributions and clip them to $[0,1]$.
    \STATE Update pairwise dominance counts across the four metrics.
  \ENDFOR
  \STATE Rank generators by descending aggregate dominance counts.
\ENDFOR
\STATE Report the metric profiles, screening outcomes,
       and ranking.
\end{algorithmic}
\end{algorithm}

\paragraph{Exchangeability summary.}
In addition to the four diagnostic metrics, we report a
composite score for complete profiles:
\[
\Emin=
\min\!\left\{
\frac{\Q+\U}{2},
\frac{\Ind+\Simi}{2}
\right\}.
\]

The first average summarizes the predictive properties of
generated samples, while the second summarizes the alignment
between real and generated samples in the chosen representation.
Taking the minimum limits compensation between these two
aspects of evaluation. We refer to this composite as \textit{Exchangeability}
in the experimental results. It provides a compact summary
of the diagnostic profile but is not included as an additional
metric in the pairwise dominance comparisons, since it is
derived from the four component metrics. When normalized
Utility is unavailable, Exchangeability is also withheld.

\FloatBarrier

\Needspace{7\baselineskip}

\section{Relation to Existing Evaluation Methods}
\label{sec:related}

Existing approaches to generative model evaluation address
complementary questions about sample fidelity, distributional
coverage, diversity, and downstream predictive performance.
Distributional measures such as FID, KID, and
MMD~\citep{heusel2017gans,binkowski2018demystifying,gretton2012kernel},
precision--recall and
density--coverage~\citep{sajjadi2018assessing,kynkaanniemi2019improved,naeem2020reliable},
and more recent approaches such as Feature Likelihood Divergence
and PQMass~\citep{jiralerspong2023feature,lemos2024pqmass}
characterize different aspects of the relationship between real
and generated distributions. Vendi complements these measures
by quantifying within-set diversity through a similarity
kernel~\citep{friedman2023vendi}.
These approaches are complementary to the task-based comparisons
in \method{}.

The closest methodological connections are to classifier-based
evaluation protocols, which assess generated data through
downstream predictive performance
\citep{shmelkov2018good,hyland2017real,yoon2019timegan,xu2019ctgan}.
\method{} distinguishes predictive performance when training
exclusively on generated samples (\textit{Quality}) from the
incremental predictive gain relative to a matched addition of
real samples (\textit{Utility}).
\textit{Indistinguishability} follows the classifier two-sample
test approach~\citep{lopezpaz2016revisiting}, while
\textit{Similarity} examines local mixing between real and
generated samples within task classes. Consequently, we do
not treat classifier two-sample testing as an independent
baseline for Indistinguishability.

Utility also addresses a different question from authenticity
and memorization audits
\citep{meehan2020non,alaa2022faithful,carlini2023extracting}.
Rather than determining whether individual samples reproduce
training examples, it measures their incremental predictive
contribution when added to existing real training data.
These forms of evaluation can be used together to investigate
whether limited augmentation gains coincide with copying.

The methodological contribution of \method{} lies in combining
these complementary evaluation perspectives with conditional
finite-sample bounds and a ranking procedure that uses
metric-specific acceptance screens and Monte Carlo comparisons
based on variability across resamples. Rather than reducing
generative performance to a single empirical score, the protocol
reports which evaluation criteria are satisfied and which
properties account for differences between generators.
Appendix~\ref{app:standard-metric-comparison} examines the
behavior of \method{} and several existing metrics under
matched controlled perturbations.

\FloatBarrier
\FloatBarrier

\section{Empirical Evaluation}
\label{sec:empirical}

The controlled experiments introduce known changes to the generated samples and test whether the probes respond as predicted. A separate comparison of two augmentation rules examines whether similar generated-only performance accompanies similar gains. We then apply the protocol to learned generators for molecules, images, and text, with three probes in the text experiment.

\subsection{Controlled Failure Experiments}
\label{sec:synthetic-quantile}

We construct a balanced binary classification problem and form four equal, stratified splits: $\DtrainOne$, $\DtrainTwo$, $\Dgen$, and $\Dtest$. All four start from the same task distribution. In the matched comparison, we alter $\Dgen$ and hold the real splits fixed. The separate class-imbalance sweep changes the class proportions in every split, including $\Dtest$, while keeping split sizes fixed. Appendix~\ref{appendix:perturbation_descriptions} describes the perturbations.
Exact copying replaces a fraction of $\Dgen$ with examples from $\DtrainOne$. Noisy copying perturbs the copied features. For mode collapse, we replace a fraction of the generated samples in each class with copies of one representative from that class. Label flipping changes the labels while holding the features fixed. The feature-noise experiment adds Gaussian noise to the generated features.At each perturbation level, we compute the four diagnostic metrics
following the evaluation protocol described in Section~\ref{sec:method}.

We specify the expected responses before inspecting the results. Exact
copies preserve their feature--label pairs, so we expect little change
in Quality and a decline in Utility as copies repeat examples available
to the learner. We also expect noisy copying to reduce Utility, with its
effect on Quality depending on the amount of noise. Within-class collapse
can lower Utility and local mixing. Its effect on Quality depends on the
task information retained by the class representatives. We expect label
flipping to reduce Quality and Utility and potentially weaken within-class
mixing. Indistinguishability and the other feature-only scores should stay
stable when the features and resampling indices are fixed. We expect
Gaussian feature noise to reduce Quality and Utility as the features
become less informative. The matched comparison tracks how these changes
also affect sample geometry and diversity
(Appendix~\ref{app:standard-metric-comparison}).

Table~\ref{tab:spearman-compact} reports the separate sensitivity sweep over
dataset sizes, classifiers, scorers, and class ratios
(Appendix~\ref{appendix:Experiments}).

\begin{table}[!htbp]
\centering
\scriptsize
\renewcommand{\arraystretch}{1}
\setlength{\tabcolsep}{3.5pt}
\caption{One-factor sensitivity analysis over dataset size, classifier, scorer, and positive-class ratio. Spearman correlations use the negative of each score, so negative correlations indicate that the score increases with severity. The matched comparison uses a fixed evaluation configuration (Appendix~\ref{app:standard-metric-comparison}).}
\label{tab:spearman-compact}
\begin{tabularx}{\linewidth}{l l *{5}{>{\centering\arraybackslash}X}}
\toprule
\textbf{Perturbation} & \textbf{Variation} & \textbf{Quality} & \textbf{Utility} & \textbf{Similarity} & \textbf{Indist.} & \textbf{Exchange.} \\
\midrule
\textbf{Exact Copies} & Dataset Size &
\cellcolor{cyan!2}{0.1083} & \cellcolor{cyan!18}{0.8915} & \cellcolor{red!20}{-0.9923} & \cellcolor{red!19}{-0.9474} & \cellcolor{cyan!14}{0.7184} \\
& Classifier &
\cellcolor{red!7}{-0.3349} & \cellcolor{cyan!9}{0.4261} & \cellcolor{red!20}{-0.9923} & \cellcolor{red!19}{-0.9281} & \cellcolor{red!3}{-0.1549} \\
& Scorer &
\cellcolor{cyan!2}{0.0469} & \cellcolor{cyan!18}{0.9923} & \cellcolor{red!20}{-0.9923} & \cellcolor{red!19}{-0.9735} & \cellcolor{cyan!18}{0.9762} \\
& Positive Ratio &
\cellcolor{red!6}{-0.3238} & \cellcolor{cyan!12}{0.6245} & \cellcolor{red!20}{-0.9922} & \cellcolor{red!20}{-0.9894} & \cellcolor{red!4}{-0.1810} \\
& \textbf{Default} &
\cellcolor{gray!5}{\texttt{--}} &
\cellcolor{cyan!20}{\textbf{1.0000}} &
\cellcolor{red!20}{\textbf{-1.0000}} &
\cellcolor{red!19}{\textbf{-0.9701}} &
\cellcolor{cyan!20}{\textbf{0.9762}} \\
\midrule
\textbf{Mode Collapse} & Dataset Size &
\cellcolor{cyan!16}{0.7953} & \cellcolor{cyan!16}{0.8100} & \cellcolor{cyan!20}{0.9923} & \cellcolor{cyan!20}{0.9817} & \cellcolor{cyan!19}{0.9619} \\
& Classifier &
\cellcolor{cyan!18}{0.9065} & \cellcolor{cyan!18}{0.8844} & \cellcolor{cyan!20}{0.9923} & \cellcolor{cyan!20}{0.9923} & \cellcolor{cyan!20}{0.9896} \\
& Scorer &
\cellcolor{cyan!19}{0.9659} & \cellcolor{cyan!20}{0.9923} & \cellcolor{cyan!20}{0.9923} & \cellcolor{cyan!20}{0.9923} & \cellcolor{cyan!20}{0.9923} \\
& Positive Ratio &
\cellcolor{cyan!16}{0.7956} & \cellcolor{cyan!17}{0.8646} & \cellcolor{cyan!20}{0.9898} & \cellcolor{cyan!20}{0.9922} & \cellcolor{cyan!19}{0.9571} \\
& \textbf{Default} &
\cellcolor{cyan!20}{\textbf{1.0000}} & \cellcolor{cyan!20}{\textbf{1.0000}} & \cellcolor{cyan!20}{\textbf{1.0000}} & \cellcolor{cyan!20}{\textbf{1.0000}} & \cellcolor{cyan!20}{\textbf{1.0000}} \\
\midrule
\textbf{Noisy Copies} & Dataset Size &
\cellcolor{red!7}{-0.3268} & \cellcolor{cyan!17}{0.8438} & \cellcolor{red!20}{-0.9889} & \cellcolor{red!19}{-0.9359} & \cellcolor{cyan!11}{0.5627} \\
& Classifier &
\cellcolor{red!8}{-0.4106} & \cellcolor{cyan!8}{0.3975} & \cellcolor{red!20}{-0.9923} & \cellcolor{red!18}{-0.9105} & \cellcolor{red!3}{-0.1391} \\
& Scorer &
\cellcolor{red!4}{-0.0391} & \cellcolor{cyan!17}{0.9899} & \cellcolor{red!20}{-0.9923} & \cellcolor{red!18}{-0.9681} & \cellcolor{cyan!17}{0.9885} \\
& Positive Ratio &
\cellcolor{red!7}{-0.3474} & \cellcolor{cyan!11}{0.5376} & \cellcolor{red!20}{-0.9922} & \cellcolor{red!20}{-0.9918} & \cellcolor{red!4}{-0.1851} \\
& \textbf{Default} &
\cellcolor{gray!5}{\texttt{--}} &
\cellcolor{cyan!20}{\textbf{1.0000}} &
\cellcolor{red!20}{\textbf{-1.0000}} &
\cellcolor{red!20}{\textbf{-1.0000}} &
\cellcolor{cyan!20}{\textbf{1.0000}} \\
\midrule
\textbf{Flipping Labels} & Dataset Size &
\cellcolor{cyan!19}{0.9383} & \cellcolor{cyan!14}{0.6937} & \cellcolor{cyan!19}{0.9643} & \cellcolor{gray!5}{0.0513} & \cellcolor{cyan!19}{0.9494} \\
& Classifier &
\cellcolor{cyan!20}{0.9880} & \cellcolor{cyan!16}{0.7779} & \cellcolor{cyan!20}{0.9923} & \cellcolor{gray!5}{-0.0866} & \cellcolor{cyan!19}{0.9659} \\
& Scorer &
\cellcolor{cyan!19}{0.9798} & \cellcolor{cyan!16}{0.7579} & \cellcolor{cyan!20}{0.9923} & \cellcolor{cyan!7}{0.2585} & \cellcolor{cyan!20}{0.9923} \\
& Positive Ratio &
\cellcolor{cyan!19}{0.9652} & \cellcolor{cyan!15}{0.7371} & \cellcolor{cyan!20}{0.9886} & \cellcolor{gray!5}{-0.0709} & \cellcolor{cyan!20}{0.9852} \\
& \textbf{Default} &
\cellcolor{cyan!20}{\textbf{1.0000}} &
\cellcolor{cyan!15}{\textbf{0.7638}} &
\cellcolor{cyan!20}{\textbf{1.0000}} &
\cellcolor{gray!2}{\textbf{0.0127}} &
\cellcolor{cyan!20}{\textbf{1.0000}} \\
\midrule
\textbf{Gaussian Noise} & Dataset Size &
\cellcolor{cyan!18}{0.9214} & \cellcolor{cyan!16}{0.7950} & \cellcolor{cyan!19}{0.9735} & \cellcolor{cyan!12}{0.5974} & \cellcolor{cyan!18}{0.8964} \\
& Classifier &
\cellcolor{cyan!17}{0.8449} & \cellcolor{cyan!12}{0.5967} & \cellcolor{cyan!20}{0.9923} & \cellcolor{cyan!20}{0.9915} & \cellcolor{cyan!14}{0.7219} \\
& Scorer &
\cellcolor{cyan!20}{0.9923} & \cellcolor{cyan!17}{0.8472} & \cellcolor{cyan!20}{0.9923} & \cellcolor{cyan!20}{0.9897} & \cellcolor{cyan!20}{0.9923} \\
& Positive Ratio &
\cellcolor{cyan!18}{0.9109} & \cellcolor{cyan!17}{0.8274} & \cellcolor{cyan!18}{0.9087} & \cellcolor{cyan!20}{0.9872} & \cellcolor{cyan!19}{0.9547} \\
& \textbf{Default} &
\cellcolor{cyan!20}{\textbf{1.0000}} &
\cellcolor{cyan!17}{\textbf{0.8729}} &
\cellcolor{cyan!20}{\textbf{1.0000}} &
\cellcolor{cyan!20}{\textbf{1.0000}} &
\cellcolor{cyan!20}{\textbf{1.0000}} \\
\bottomrule
\end{tabularx}
\end{table}

As exact or noisy copying increases, Utility consistently declines
across the reported settings, whereas Similarity and
Indistinguishability increase. This explains their negative
Spearman correlations under the degradation-oriented convention:
although copying reduces the incremental predictive value of
generated samples, it can improve their resemblance to real
training data. Similar behavior is observed for several
conventional distributional metrics
(Appendix~\ref{app:standard-metric-comparison}).
The matched comparison confirms the increase in Similarity,
although the response of Indistinguishability differs across
experimental configurations. These results highlight the
importance of distinguishing distributional resemblance from
incremental predictive value when evaluating generative models.

Mode collapse and Gaussian feature noise reduce all four scores across
the settings in Table~\ref{tab:spearman-compact}. Under label flipping,
Quality, Utility, and Similarity decline. The Indistinguishability
correlations are closer to zero and vary in sign.

\FloatBarrier

\subsection{Utility and Informative Augmentation}
\label{sec:utility-isolation}

To further investigate the distinction between Quality and Utility,
we compare two generation strategies that preserve class-related
structure but explore the feature space differently.
SMOTE generates samples by interpolating between existing
same-class observations~\citep{chawla2002smote}, whereas
Transferred Local Differences (TLD) generates samples of the form
$A+\alpha(C-B)$ by transferring the displacement between two
same-class neighbors $B$ and $C$ to an anchor $A$ from the same class.
Unlike interpolation, this construction allows generated samples
to extend beyond the line segments connecting existing observations.
\begin{figure}[!htbp]
  \centering
  \begin{subfigure}[t]{0.56\linewidth}
    \centering
    \includegraphics[width=\linewidth]{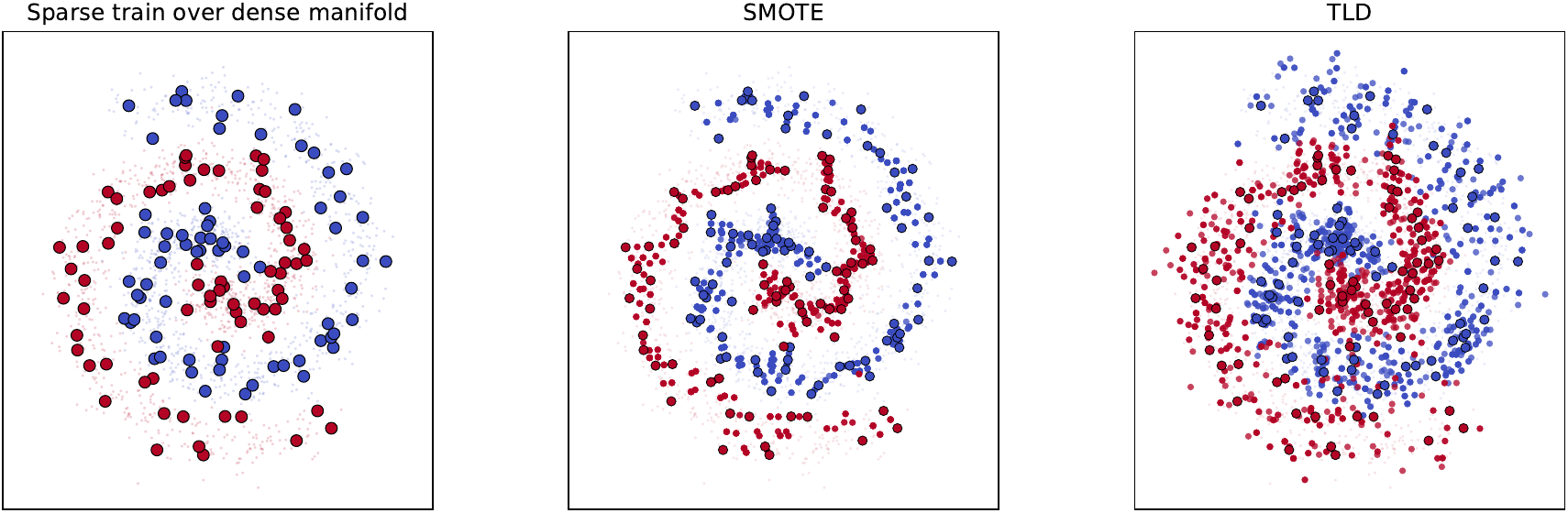}
    \caption{Two spirals: sparse training set, SMOTE, and TLD.}
  \end{subfigure}\hfill
  \begin{subfigure}[t]{0.41\linewidth}
    \centering
    \includegraphics[width=\linewidth]{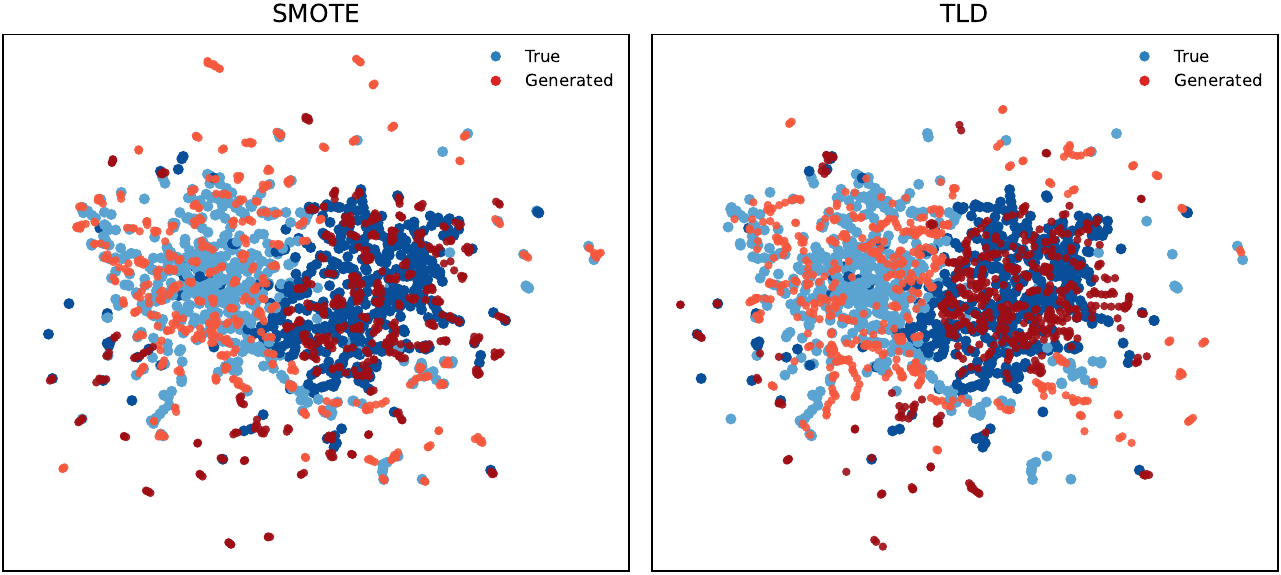}
    \caption{MNIST projection: SMOTE and TLD.}
  \end{subfigure}
  \caption{Comparison of interpolation-based generation (SMOTE)
  and generation through transferred local differences (TLD).
  The two-spirals panels illustrate how the two strategies
  populate the feature space from a sparse real training set.
  The MNIST panels show a UMAP visualization of real and
  generated samples.}
  \label{fig:utility-diagnostic}

  \medskip
  \begin{minipage}{\linewidth}
    \centering
    \setlength{\tabcolsep}{15pt}
    \captionof{table}{Quality, normalized Utility, and
    unnormalized augmentation gain for SMOTE and TLD.
    Values are mean $\pm$ standard deviation across random seeds.}
    \label{tab:utility-diagnostic}
    \small
    \begin{adjustbox}{max width=\linewidth}
    \begin{tabular}{@{}llccc@{}}
    \toprule
    \textbf{Dataset} &
    \textbf{Generator} &
    \textbf{Quality} &
    \textbf{Utility} &
    \textbf{Utility Gain} \\
    \midrule
    Two spirals & SMOTE &
    $1.000 \pm 0.000$ &
    $0.409 \pm 0.150$ &
    $0.014 \pm 0.003$ \\

    Two spirals & TLD &
    $1.000 \pm 0.000$ &
    $\mathbf{0.728 \pm 0.099}$ &
    $\mathbf{0.026 \pm 0.010}$ \\
    \midrule
    MNIST latent & SMOTE &
    $0.983 \pm 0.011$ &
    $0.027 \pm 0.031$ &
    $0.0002 \pm 0.0019$ \\

    MNIST latent & TLD &
    $0.989 \pm 0.008$ &
    $\mathbf{0.093 \pm 0.054}$ &
    $\mathbf{0.0029 \pm 0.0016}$ \\
    \bottomrule
    \end{tabular}
    \end{adjustbox}
  \end{minipage}
\end{figure}

We evaluate both strategies on the two-spirals problem and on
MNIST digits 3 and 8. For MNIST, generation is performed in a
six-dimensional projection, with UMAP used for
visualization~\citep{mcinnes2018umap}.
Figure~\ref{fig:utility-diagnostic} illustrates the different
sampling patterns produced by the two strategies.

Both generation strategies achieve comparable Quality on each
task (Table~\ref{tab:utility-diagnostic}), indicating that their
generated samples support similar predictive performance when
used independently. However, TLD achieves higher mean Utility
than SMOTE on both two spirals ($0.728$ vs.\ $0.409$) and MNIST
($0.093$ vs.\ $0.027$), together with larger unnormalized
augmentation gains.

These results demonstrate why generated-only predictive performance
and incremental predictive value should be evaluated separately.
Although both strategies preserve task-relevant structure to a
similar extent, they differ in how much they improve learning from
existing real data. Quality alone does not reveal this difference,
whereas Utility explicitly captures the additional predictive
contribution of the generated samples.

\FloatBarrier

\subsection{Evaluation of Learned Generators}
\label{sec:supporting-real}

\subsubsection{Molecular Graphs}
\label{sec:real-world-graphs}

We evaluate molecular generators on five classification datasets
from the Therapeutics Data Commons~\citep{huang2021tdc}, covering
Ames mutagenicity, blood--brain barrier penetration, CYP1A2 and
CYP2C19 inhibition, and hERG blockade. The dataset characteristics
and split sizes are provided in
Appendix~\ref{appendix:molecular_datasets}.

Molecules are represented using molecular fingerprints and
graph-based feature representations
(Appendix~\ref{appendix:Vectorizers}).
Generated molecules receive the label used for conditioning
or, for class-specific generators, the label of the training class.

We compare several families of generative models, including
STGG~\citep{ahn2021spanning},
WGAN-GP with an R-GCN discriminator~\citep{keras2021wgan},
JTNN~\citep{jin2018junction},
HierVAE~\citep{jin2020hiervae},
MoFlow~\citep{zang2020moflow},
GDSS~\citep{jo2022score},
SwinGNN~\citep{yan2023swingnn},
the S4DD chemical language model~\citep{ozcelik2024s4},
and three variants of a Neighborhood Swap baseline
that we introduce for comparison.
The generator architectures, training configurations, and
pretraining strategies are detailed in
Appendix~\ref{app:molecular_generators}.

Figure~\ref{fig:generators-rank} summarizes the performance
of the generators across the five prediction tasks and
the different feature representations.

\begin{figure}[!htbp]
  \centering
  \includegraphics[width=0.95\linewidth]{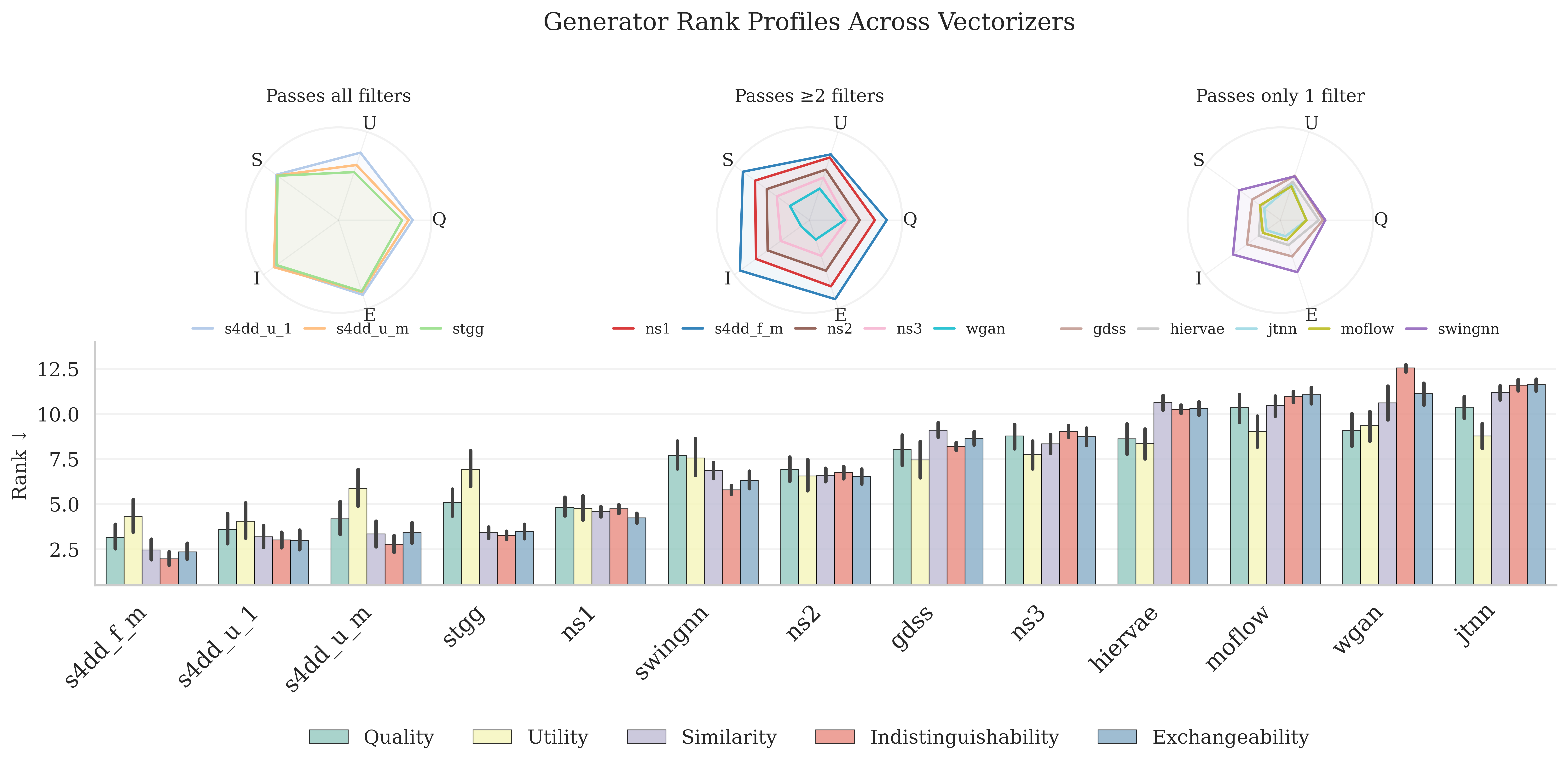}
  \caption{Evaluation of molecular graph generators across
  prediction tasks and feature representations.
  The radar panels group generators according to the number
  of acceptance screens they pass
  (Table~\ref{tab:bounds_check}).
  The bar plot reports mean ranks for the four diagnostic
  metrics and the Exchangeability summary, with lower
  ranks indicating better performance.}
  \label{fig:generators-rank}
\end{figure}

The S4DD variants and STGG achieve favorable ranks across
several metrics, although their relative performance varies
with the evaluation criterion. STGG, for instance, ranks
better on Similarity and Indistinguishability than on Utility,
indicating that its favorable alignment with real samples
is not accompanied by an equally strong contribution to
downstream learning.

The screening outcomes provide an additional perspective.
Although the filtered S4DD variant achieves favorable mean
ranks, it does not satisfy all acceptance criteria.
Conversely, the two unfiltered S4DD configurations and STGG
pass all four metric screens. This distinction illustrates
why relative metric performance and satisfaction of prescribed
acceptance criteria are considered separately in the
\method{} ranking procedure.

\FloatBarrier
\subsubsection{Images}
\label{sec:image-benchmarks}

\begin{figure}[!htbp]
  \centering
  \includegraphics[width=0.95\linewidth]{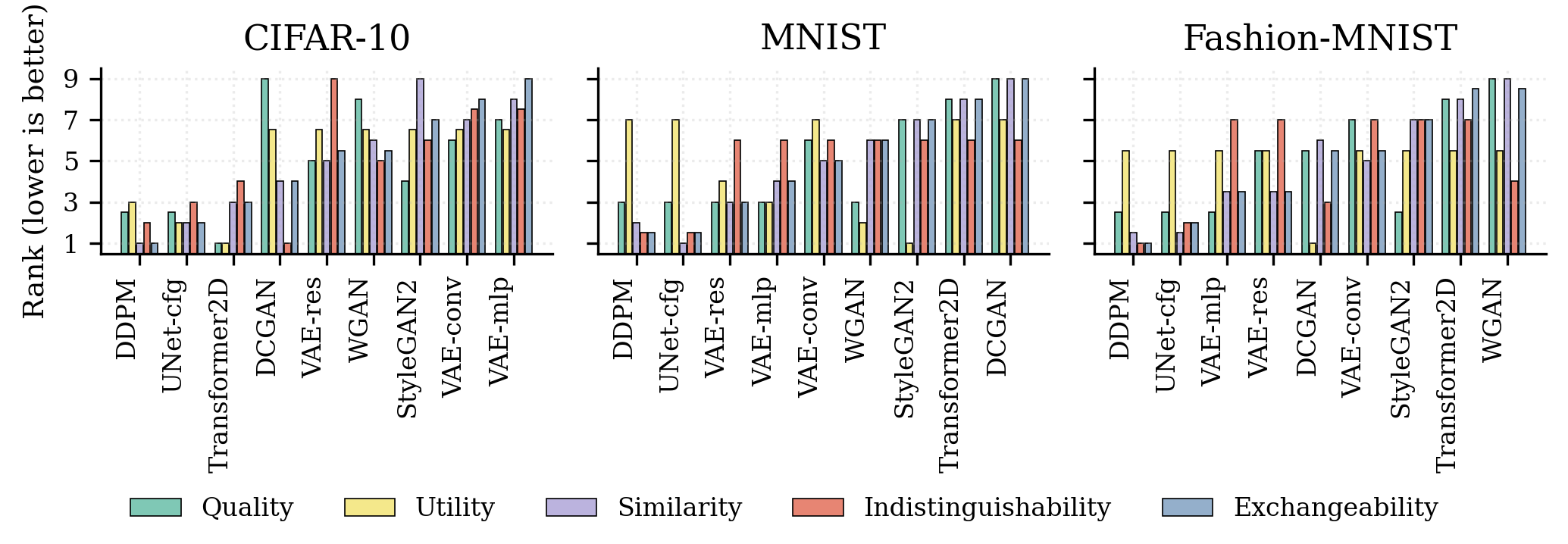}
  \caption{Metric-specific ranks of image generators on
  CIFAR-10, MNIST, and Fashion-MNIST, together with their
  Exchangeability ranks. Lower ranks indicate better
  performance under the corresponding metric.}
  \label{fig:cifar10-ranks}
\end{figure}

We evaluate VAEs, GANs, and diffusion models on
MNIST~\citep{lecun1998gradient},
Fashion-MNIST~\citep{xiao2017fashion}, and
CIFAR-10~\citep{krizhevsky2009learning}, using the class labels
provided by each dataset to define the prediction task.
Figure~\ref{fig:cifar10-ranks} presents the generator ranks
for each \method{} metric. Appendix~\ref{app:image-results}
provides the corresponding metric values and comparisons
with FID~\citep{heusel2017gans},
KID~\citep{binkowski2018demystifying},
precision and recall, Inception Score, and kernel-based
diversity measures.

On CIFAR-10, DDPM and UNet-CFG achieve favorable ranks
across both conventional distributional metrics and several
\method{} metrics. Transformer2D achieves the highest
Quality and Utility, despite ranking fourth on FID and KID.
Conversely, StyleGAN2-lite ranks third on both FID and KID,
but achieves a Utility of $0.000$ and the lowest Similarity
among the evaluated generators
(Tables~\ref{tab:cifar10-ranked}
and~\ref{tab:vals-ranks-cifar10}).

The relative performance of generators also varies across
datasets. On MNIST, DDPM and UNet-CFG achieve high Quality
and comparatively high Similarity and Indistinguishability,
whereas StyleGAN2-lite and WGAN-GP achieve the highest
Utility scores (Table~\ref{tab:vals-ranks-mnist}).
On Fashion-MNIST, several generators achieve comparable
Quality, but only DCGAN exhibits positive Utility at the
reported precision (Table~\ref{tab:vals-ranks-fashionmnist}).

These comparisons illustrate that the relative strengths
of generative models depend on both the evaluation criterion
and the downstream task. In particular, favorable distributional
scores and strong generated-only predictive performance do
not necessarily translate into gains from data augmentation.

\FloatBarrier
% \Needspace{0.25\textheight}
\subsubsection{Text}
\label{subsec:text}

To demonstrate the applicability of \method{} to text generation,
we use the AG-News topic categories~\citep{zhang2015character}
to condition headline generation with DistilGPT-2, GPT-2,
and GPT-2 Medium~\citep{radford2019gpt2,distilgpt2modelcard}.
Headlines are embedded using MiniLM~\citep{wang2020minilm},
and logistic regression is used as the task classifier
for topic prediction.

\begin{table}[H]
\centering
\small
\setlength{\tabcolsep}{8pt}
\caption{Evaluation of text generators on AG-News using
generated-only classification performance and three
\method{} diagnostic metrics.}
\label{tab:text-results}
\begin{tabular}{lcccc}
\toprule
\textbf{Model} & \textbf{F1} & $\Q$ & $\Ind$ & $\Simi$ \\
\midrule
\texttt{distilgpt2}  & 0.38 & 0.48 & 0.04 & 0.17 \\
\texttt{gpt2}        & 0.45 & 0.57 & 0.08 & 0.24 \\
\texttt{gpt2-medium} & 0.45 & 0.57 & 0.09 & 0.26 \\
\bottomrule
\end{tabular}
\end{table}

As shown in Table~\ref{tab:text-results}, GPT-2 and GPT-2 Medium
achieve higher Quality than DistilGPT-2, indicating better
preservation of task-relevant information. Although the two
GPT-2 configurations achieve identical Quality at the reported
precision, GPT-2 Medium exhibits slightly higher Similarity
and Indistinguishability. Nevertheless, the low values of
Indistinguishability across all three models indicate that
generated headlines remain readily distinguishable from real
headlines in the selected representation.

This experiment illustrates the application of \method{} to
text through a task-specific feature representation and
classifier. Utility is not reported because the matched
real-data augmentation comparison is unavailable in this setup.

\FloatBarrier

\section{Uncertainty Screens and Generator Comparisons}
\label{sec:analysis}

We examine how the uncertainty screens behave across different
classifier configurations and how their outcomes inform the
comparison of generative models.

Table~\ref{tab:pac-fail} reports the screen failure rates across
classifiers in the controlled experiments. Utility exhibits
substantial variation, with failure rates of $50.0\%$ for
Gaussian naive Bayes and $48.2\%$ for logistic regression,
compared with $0\%$ for $k$-NN and SVC.
Similarity passes its screen across all reported configurations.
These results illustrate how the evaluation configuration affects
the assessment of generated data, particularly when measuring
incremental predictive gains.

\begin{table}[!htbp]
\centering
\begin{minipage}[t]{0.43\linewidth}
\centering
\scriptsize
\setlength{\tabcolsep}{1.5pt}
\renewcommand{\arraystretch}{0.93}
\begin{tabular}{lccccc}
\toprule
\textbf{Classifier} &
$\Q$ & $\U$ & $\Ind$ & $\Simi$ & $\Emin$ \\
\midrule
AdaBoost          & 0.000 & 0.167 & 0.067 & 0.000 & 0.167 \\
Decision Tree     & 0.006 & 0.167 & 0.042 & 0.000 & 0.167 \\
Extra Trees       & 0.048 & 0.167 & 0.194 & 0.000 & 0.167 \\
GaussianNB        & 0.273 & 0.500 & 0.064 & 0.000 & 0.500 \\
Gradient Boosting & 0.027 & 0.167 & 0.085 & 0.000 & 0.167 \\
k-NN              & 0.145 & 0.000 & 0.033 & 0.000 & 0.000 \\
Logistic Reg.     & 0.161 & 0.482 & 0.030 & 0.000 & 0.482 \\
Random Forest     & 0.048 & 0.167 & 0.124 & 0.000 & 0.167 \\
SVC               & 0.215 & 0.000 & 0.048 & 0.000 & 0.000 \\
\bottomrule
\end{tabular}
\caption{Fraction of controlled experimental runs that fail
each uncertainty screen, using $\delta=0.05$.}
\label{tab:pac-fail}
\end{minipage}\hfill
\begin{minipage}[t]{0.53\linewidth}
\centering
\scriptsize
\renewcommand{\arraystretch}{0.85}
\setlength{\tabcolsep}{6pt}
\begin{tabular}{lccccc}
\toprule
\textbf{Generator} &
$\Q$ & $\U$ & $\Ind$ & $\Simi$ & $\Emin$ \\
\midrule

s4dd\_u\_1
& \cellcolor{passblue}\cmark
& \cellcolor{passblue}\cmark
& \cellcolor{passblue}\cmark
& \cellcolor{passblue}\cmark
& \cellcolor{passblue}\cmark \\

s4dd\_u\_m
& \cellcolor{passblue}\cmark
& \cellcolor{passblue}\cmark
& \cellcolor{passblue}\cmark
& \cellcolor{passblue}\cmark
& \cellcolor{passblue}\cmark \\

stgg
& \cellcolor{passblue}\cmark
& \cellcolor{passblue}\cmark
& \cellcolor{passblue}\cmark
& \cellcolor{passblue}\cmark
& \cellcolor{passblue}\cmark \\

\midrule

ns1
& \cellcolor{passblue}\cmark
& \cellcolor{failred}\xmark
& \cellcolor{passblue}\cmark
& \cellcolor{passblue}\cmark
& \cellcolor{failred}\xmark \\

\midrule

s4dd\_f\_m
& \cellcolor{failred}\xmark
& \cellcolor{failred}\xmark
& \cellcolor{passblue}\cmark
& \cellcolor{passblue}\cmark
& \cellcolor{failred}\xmark \\

ns2
& \cellcolor{passblue}\cmark
& \cellcolor{failred}\xmark
& \cellcolor{passblue}\cmark
& \cellcolor{failred}\xmark
& \cellcolor{failred}\xmark \\

ns3
& \cellcolor{passblue}\cmark
& \cellcolor{failred}\xmark
& \cellcolor{passblue}\cmark
& \cellcolor{failred}\xmark
& \cellcolor{failred}\xmark \\

wgan
& \cellcolor{passblue}\cmark
& \cellcolor{failred}\xmark
& \cellcolor{passblue}\cmark
& \cellcolor{failred}\xmark
& \cellcolor{failred}\xmark \\

\midrule

gdss
& \cellcolor{failred}\xmark
& \cellcolor{failred}\xmark
& \cellcolor{passblue}\cmark
& \cellcolor{failred}\xmark
& \cellcolor{failred}\xmark \\

hiervae
& \cellcolor{failred}\xmark
& \cellcolor{failred}\xmark
& \cellcolor{passblue}\cmark
& \cellcolor{failred}\xmark
& \cellcolor{failred}\xmark \\

jtnn
& \cellcolor{failred}\xmark
& \cellcolor{failred}\xmark
& \cellcolor{passblue}\cmark
& \cellcolor{failred}\xmark
& \cellcolor{failred}\xmark \\

moflow
& \cellcolor{failred}\xmark
& \cellcolor{failred}\xmark
& \cellcolor{passblue}\cmark
& \cellcolor{failred}\xmark
& \cellcolor{failred}\xmark \\

swingnn
& \cellcolor{failred}\xmark
& \cellcolor{failred}\xmark
& \cellcolor{passblue}\cmark
& \cellcolor{failred}\xmark
& \cellcolor{failred}\xmark \\

\bottomrule
\end{tabular}
\caption{Uncertainty-screen outcomes for molecular generators,
grouped by the number of failed screens across the four
diagnostic metrics. Exchangeability is reported separately
as a composite summary.}
\label{tab:bounds_check}
\end{minipage}
\end{table}

Table~\ref{tab:bounds_check} illustrates how the screening
outcomes differentiate molecular generators. Although all
generators pass the Indistinguishability screen, only the
two unfiltered S4DD configurations and STGG pass the Utility
screen and satisfy all four metric-specific acceptance criteria.

The Neighborhood Swap baseline \texttt{ns1}, for instance,
satisfies the Quality, Indistinguishability, and Similarity
criteria but falls short on Utility. This distinction
illustrates that preserving task-relevant information and
resembling real samples do not necessarily imply sufficient
incremental predictive value.

The screening outcomes establish the initial grouping of
generators in Algorithm~\ref{alg:main-ranking}, while Monte
Carlo pairwise dominance accounts for variability in metric
values across evaluation resamples when determining the
ordering within each group. Together, these comparisons
provide a ranking that reflects both the satisfaction of
metric-specific acceptance criteria and the relative
performance of generators across the four diagnostic metrics.

\FloatBarrier
\section{Discussion and Conclusion}
\label{sec:discussion-conclusion}

In this work, we approached generative model evaluation as a
multidimensional problem in which both the properties of generated
samples and the variability of their evaluation matter.
We proposed \method{}, a task-grounded protocol that constructs
complementary diagnostic metrics primarily from classifier
performance comparisons. This formulation allows us to draw on
established theoretical results to derive conditional finite-sample
bounds and to incorporate both metric-specific acceptance criteria
and variability across resamples into the ranking of generative models.

Our experiments illustrate the importance of evaluating different
properties of generated data separately. In the controlled copying
experiments, increasing the proportion of training-set copies
reduced augmentation gains while increasing local mixing between
real and generated samples. The comparison between SMOTE and TLD
further demonstrated that similar Quality scores can accompany
different levels of Utility. Together, these results highlight
the distinction between preserving task-relevant structure,
resembling real data, and contributing additional predictive
value. A favorable score along one of these dimensions does
not necessarily imply favorable performance along the others.

This distinction also becomes important when ranking learned
generators. In the molecular benchmarks, for example, STGG
performs better in terms of Similarity and Indistinguishability
than in terms of Utility, while the Neighborhood Swap baseline
passes three of the four metric screens but falls short on Utility.
Rather than obscuring such differences through a single aggregate
score, \method{} retains the individual metric profiles.
The screening outcomes establish an initial grouping of generators,
while Monte Carlo comparisons incorporate the variability observed
across evaluation resamples. The resulting ordering can therefore
be interpreted alongside the particular strengths and weaknesses
of each generator.

\paragraph{Scope and limitations.}
\label{sec:limitations}
GUIDE evaluates generated data under a specified task and evaluation
setup, so its metric profiles describe performance in that setting
rather than intrinsic properties of a generator. Changing the feature
representation, classifier, or data split may change the results.
Utility illustrates this dependence: its normalized form is informative
only when adding real data yields a sufficiently large reference gain.
When that condition is not met, the unnormalized augmentation gain
should be reported instead. Since the metrics address only selected
properties of generated data, privacy audits and domain-specific
validity checks may also be needed.

The uncertainty results require similar care in interpretation. The
theoretical bounds hold under the assumptions in Appendix~\ref{app:theory}
and for the performance measures for which they are derived. In our
experiments, the screens use approximate radii with F1 scores; they are
motivated by the bounds but are not formal confidence bounds for those
scores. The Monte Carlo rankings summarize variation across evaluation
resamples under a chosen threshold and sampling model. They do not
capture variation from retraining a generator or establish probabilities
for population rankings. Future work could extend the bounds beyond the current performance measures and classification setting, including to regression tasks.

The experiments show why these distinctions matter. Training-set copies
increase local mixing even as augmentation gains decline, while SMOTE
and TLD achieve similar Quality but different Utility. Together, these
results show that task-signal preservation, incremental predictive
value, and alignment with real data describe separate properties of
generated samples. GUIDE brings these perspectives into a single
evaluation, allowing comparisons to show not only how generators rank,
but which measured properties support that ranking and under what
conditions it applies.

\paragraph{Acknowledgments.}
For the purpose of open access, the authors have applied a Creative Commons Attribution (CC BY) license to any Author Accepted Manuscript version arising from this submission.

\FloatBarrier
\bibliographystyle{tmlr}
\bibliography{reference}

\appendix

\input{appendix_theory}

\input{Appendix}

\end{document}

%% file: appendix_theory.tex
\section{Formal Definitions and Conditional Uncertainty Screens for \method}
\label{app:theory}

This appendix defines the diagnostic metrics and derives their conditional
finite-sample bounds. The task-score derivations use classification accuracy
($1$ minus the $0$--$1$ risk) and apply under the sampling and capacity
conditions stated below. The protocol allows other task scoring rules, but
the corresponding confidence bounds require separate derivations. The
experimental screens use F1 and approximate capacity inputs, as described
in Section~\ref{sec:method}; these screens are not established confidence
bounds for the F1-based scores.

\subsection{Formal Metric Definitions}
\label{app:theory-metric-defs}

\subsubsection{Notation Glossary}
\label{appendix:bounds}

\begingroup
\small
\setlength{\tabcolsep}{4pt}
\renewcommand{\arraystretch}{1.16}
\begin{longtable}{@{}p{0.25\linewidth}p{\dimexpr0.75\linewidth-2\tabcolsep\relax}@{}}
\caption{Notation for the scores and conditional bounds. Hats denote sample estimates, stars denote population counterparts, and $(i)$ indexes the evaluation split.}
\label{tab:notation-glossary}\\
\toprule
\textbf{Symbol} & \textbf{Definition} \\
\midrule
\endfirsthead
\multicolumn{2}{@{}l}{\small\itshape Table~\thetable\ (continued)}\\[2pt]
\toprule
\textbf{Symbol} & \textbf{Definition} \\
\midrule
\endhead
\midrule
\multicolumn{2}{r@{}}{\small\itshape Continued on the next page}\\
\endfoot
\bottomrule
\endlastfoot
\multicolumn{2}{@{}l}{\emph{Data and predictors}}\\
$D_{\text{train1}}^{(i)}$ & Real training data for the baseline task predictor.\\
$D_{\text{train2}}^{(i)}$ & Additional real data for the matched augmentation comparison.\\
$D_{\text{generated}}^{(i)}$ & Generated data used in split $i$.\\
$D_{\text{test}}$ & Held-out real data used to score the task predictors.\\
$D_r,\;D_g,\;D_{\text{mix}}$ & The real and generated samples in the Similarity probe and their pooled sample.\\
$N,\;N_r,\;N_g$ & Total, real, and generated sample counts in that pool.\\
$N_{\mathrm{disc}}$ & Total count in a balanced domain comparison. Indistinguishability compares baseline real data with generated data. The Utility discrepancy comparison uses additional real data and generated data.\\
$k$ & Number of non-self neighbors in the Similarity probe.\\
$\Phi_\theta$ & Task-predictor class.\\
$\Psi_\theta$ & Real-versus-generated discriminator class.\\
\midrule
\multicolumn{2}{@{}l}{\emph{Scores and gains}}\\*
$\rho_1^{(i)},\;\rho_2^{(i)}$ & Task scores of predictors trained on baseline real data and generated data, respectively.\\*
$\rho_3^{(i)},\;\rho_4^{(i)}$ & Task scores after adding real or generated data to the baseline training set, respectively.\\
$\rho_5^{(i)}$ & Balanced accuracy of the real-versus-generated discriminator on its evaluation pool.\\
$\Delta_{\mathrm{real}}^{(i)}$ & Real-data gain $\rho_3^{(i)}-\rho_1^{(i)}$.\\
$\Delta_{\mathrm{generated}}^{(i)}$ & Generated-data gain $\rho_4^{(i)}-\rho_1^{(i)}$.\\
$Q^{(i)},\;U^{(i)},\;I^{(i)},\;S^{(i)}$ & Quality, normalized Utility, Indistinguishability, and Similarity.\\
\midrule
\multicolumn{2}{@{}l}{\emph{Bound ingredients}}\\*
$P,\;Q$ & Real and generated data distributions in the domain-adaptation derivations.\\
$\lambda(P,Q)$ & Minimum sum of real and generated risks over the task-predictor class.\\
$\gamma^\star$ & Absolute difference in generated-domain risk between the two augmented predictors.\\
$\mathcal R_n,\;d$ & Empirical Rademacher complexity and a VC-dimension upper bound for the class under consideration.\\
$\delta$ & Failure probability allocated to a bound or a family of bounds.\\
\end{longtable}
\endgroup

\paragraph{Formal metric definitions.}
On split $i$, the task scores $\rho_1^{(i)},\ldots,\rho_4^{(i)}$ use the same
scoring rule and real test set. The discriminator score $\rho_5^{(i)}$ uses
the balanced real-versus-generated evaluation pool. Pooled datasets retain
each observation, including duplicates. The corresponding probe scores are
\begin{align}
Q^{(i)} &:= \mathrm{clip}_{[0,1]}\!\left(\frac{\rho_2^{(i)}}{\rho_1^{(i)}+\varepsilon}\right),\\
U^{(i)} &:= \mathrm{clip}_{[0,1]}\!\left(\frac{\Delta_{\mathrm{generated}}^{(i)}}{\Delta_{\mathrm{real}}^{(i)}}\right)
\quad\text{when }\Delta_{\mathrm{real}}^{(i)}>\tau,\\
I^{(i)} &:= 1-2\left|\rho_5^{(i)}-0.5\right|.
\end{align}
Here $\varepsilon>0$ stabilizes the Quality ratio. If the real-data gain is
at most $\tau=10^{-3}$, normalized Utility is unavailable and the reported
Utility Gain is $\Delta_{\mathrm{generated}}^{(i)}$.

For Similarity, let $c$ index task classes and let $j$ index queries within
the pooled class-$c$ sample on split $i$. Let $\hat p_{cj}$ be the fraction
of generated points among the $k$ non-self neighbors of query $j$, with
$1\le k<N_c$. Let
$H_2(p)=-p\log_2 p-(1-p)\log_2(1-p)$ with $H_2(0)=H_2(1)=0$. Then
\begin{equation}
S_c^{(i)}:=\frac{1}{N_c}\sum_{j=1}^{N_c} H_2(\hat p_{cj}),
\qquad
S^{(i)}:=\operatorname{gmean}_{c\in\mathcal C} S_c^{(i)} .
\end{equation}
Task labels determine the class pools, and the entropy is computed from
the proportion of generated samples in each neighborhood.
For a complete profile, the composite is
\begin{equation}
\mathcal E_{\min}^{(i)}
=\min\!\left\{\frac{Q^{(i)}+U^{(i)}}2,\,
              \frac{I^{(i)}+S^{(i)}}2\right\}.
\end{equation}

\subsection{PAC Assumptions and Notation}
\label{app:theory-assumptions}

\subsubsection{Bound Ingredients and Assumptions}
\label{app:bound-ingredients}

We condition on generator fitting and on the fitted predictors when
evaluation samples are independent of fitting. Any data-dependent choice
made using an evaluation pool requires its own uniform bound or an
independent pool. Where a domain-discrepancy bound is used, the relevant
real and generated samples must represent the distributions in that bound.

\paragraph{Confidence allocation.}
For simultaneous claims across $M$ generators, $K$ probes, and $R$ resamples,
one may allocate $\delta_*=\delta/(MKR)$ to each probe bound. When a probe
uses several stochastic inequalities, their failure probabilities must sum
to at most its allocated $\delta_*$. A union bound combines those events
even if resamples overlap. The resulting probability statements require
valid bounds for all inputs to each event.

\paragraph{Domain adaptation.}
Let $P$ and $Q$ denote the real and generated distributions, let
$\mathcal H$ be the task-predictor class, and write
$R_D(h)=\Pr_{(x,y)\sim D}[h(x)\ne y]$. Following \citet{ben2010theory}, define
\begin{align}
\mathrm{disc}_{\mathcal H\Delta\mathcal H}(P,Q)
&:=2\sup_{h,h'\in\mathcal H}
 \left|\Pr_P[h(x)\ne h'(x)]-\Pr_Q[h(x)\ne h'(x)]\right|,
 \label{eq:DA}\\
\lambda(P,Q)&:=\inf_{h\in\mathcal H}\bigl(R_P(h)+R_Q(h)\bigr).
\end{align}
The disagreement probabilities use the input marginals of $P$ and $Q$.
The $0$--$1$ triangle inequality gives
\begin{equation}
|R_P(h)-R_Q(h)|
\le \tfrac12\,\mathrm{disc}_{\mathcal H\Delta\mathcal H}(P,Q)+\lambda(P,Q)
\quad(h\in\mathcal H).
\label{eq:DA-inst}
\end{equation}
Indeed, for any $h'\in\mathcal H$, bound $R_P(h)$ by
$\Pr_P[h(x)\ne h'(x)]+R_P(h')$, transfer the disagreement term to $Q$, and
bound the resulting disagreement by $R_Q(h)+R_Q(h')$. Taking the infimum over
$h'$ and interchanging $P$ and $Q$ yields \eqref{eq:DA-inst}.

Let $\bar d$ and $\bar\lambda$ be valid upper bounds for the discrepancy and
joint risk. On a balanced sample, $2(2\hat a-1)$ estimates the empirical
discrepancy when $\hat a$ is the optimal domain-classification accuracy over
the disagreement indicators and their complements. An upper bound from
this estimate requires a uniform deviation bound for that class and, if
optimization is approximate, an allowance for the optimization gap. An
arbitrary fitted discriminator gives a diagnostic estimate of separability.
A valid $\bar\lambda$ can use upper confidence limits for the two risks of
a fixed candidate $h\in\mathcal H$ evaluated on independent samples, since
the sum of its risks upper-bounds the infimum. The empirical minimum alone
does not provide this coverage. The deterministic upper bounds
$\bar d=2$ and $\bar\lambda=2$ are always valid.

\subsubsection{Practical Capacity Choices for Conditional Bounds}
\label{app:bound-implementation}

\paragraph{Task-classifier complexity.}
Uniform bounds based on the empirical Rademacher complexity of the loss
class are available under independent sampling
\citep{bartlett2002rademacher}. Use the convention
$\mathcal R_n=\mathbb E_{\sigma}\sup_{f\in\mathcal F}
n^{-1}\sum_{i=1}^{n}\sigma_i f(z_i)$ on the fixed evaluation points $z_i$,
where $\mathcal F$ is the loss class and the signs $\sigma_i$ are independent
and uniform on $\{-1,1\}$. For binary classification with a valid
VC-dimension upper bound $1\le d\le n$, Sauer--Shelah and the finite-class
Rademacher bound give
\[
\mathcal R_n\le \sqrt{\frac{2d\log(en/d)}{n}} .
\]
The experimental screens use $\sqrt{2/n}$ when a justified capacity bound
is unavailable. Some synthetic settings use classifier-specific surrogates
in the range $0.01$--$0.09$. These quantities are screen inputs whose values
can affect the pass/fail results.

\paragraph{Discriminator capacity and risk inputs.}
The experimental screens use $d_{\mathrm{disc}}=50$ as a capacity surrogate.
A VC confidence statement requires a justified bound for the class in the
particular calculation. The disagreement class in \eqref{eq:DA} and the
real-versus-generated discriminator in Indistinguishability need their own capacity
controls. The Utility screen can use
empirical estimates of $\lambda$ and $\gamma^\star$. The worst-case inputs
are $2$ and $1$, respectively, when those estimates are unavailable.
The documented Utility diagnostic divides the plug-in error term
\[
\widehat{\mathrm{disc}}+\widetilde\varepsilon_{\mathrm{disc}}
+\hat\gamma+6\widetilde\varepsilon_{\mathrm{gen}}+2\hat\lambda
\]
by the observed real-data gain when it exceeds $\tau$. Here
$\widetilde\varepsilon_{\mathrm{gen}}=\sqrt{\log(6/\delta)/(2N_{\min})}$
uses the minimum of the three training-sample sizes and the real test size,
and $\widetilde\varepsilon_{\mathrm{disc}}
=\sqrt{[4d_{\mathrm{disc}}\log(2N_{\mathrm{disc}})+\log(8/\delta)]/N_{\mathrm{disc}}}$.
These are the plug-in terms for interpreting the reported screens. The
confidence bound uses the justified upper limits in
Appendix~\ref{app:theory-utility}.

\paragraph{Local entropy sensitivity.}
For the binary entropy $H_2$ used in Similarity, the derivative is
$H_2'(p)=\log_2((1-p)/p)$. Its absolute value on the interior interval
$[\alpha,1-\alpha]$, with $\alpha=1/(k+1)$, is bounded by
$L_{H_2}=\log_2((1-\alpha)/\alpha)=\log_2 k$. An entropy radius computed
using natural logarithms is in nats and is divided by $\ln 2$ to express it
in bits. This slope bound applies when both proportions lie in the indicated
interval. The argument in Appendix~\ref{app:similarity-bound} handles the
endpoints directly.

\paragraph{Quantile-to-moment mapping.}
\label{app:quantile-to-moment}
For the Monte Carlo ranking procedure, we convert the empirical quartiles
$(Q_1,M,Q_3)$ into the inputs
\[
\hat{\mu}_Q=\frac{Q_1+M+Q_3}{3},
\qquad
\hat{\sigma}_Q=\frac{\mathrm{IQR}}{1.34898}.
\]
The mean approximation follows \citet{wan2016estimatingsamplemeanstandard}.
The scale uses the large-sample normal approximation to their
sample-size-dependent denominator. The reported summaries remain the
median and IQR.

\subsection{Quality Bound}
\label{app:theory-quality}
\label{appendix:quality-bound}

Let $h_g$ and $h_r$ be trained on generated and baseline real data,
respectively. Write $a_g^\star=1-R_P(h_g)$ and $a_r^\star=1-R_P(h_r)$
for their real-domain accuracies. Let $\hat R_g(h_g)$ be the risk on an
independent generated hold-out and $\hat a_r$ the accuracy on the real
test set. Suppose that the joint event includes
\[
R_Q(h_g)\le \hat R_g(h_g)+e_g,\qquad a_r^\star\le \hat a_r+e_r ,
\]
as well as the bounds $\bar d$ and $\bar\lambda$ from
Appendix~\ref{app:bound-ingredients}. For predictors fixed independently of
their evaluation pools, $e_g$ and $e_r$ can be obtained from bounded-loss
concentration. By \eqref{eq:DA-inst},
\begin{align}
L_g&:=\max\{0,1-\hat R_g(h_g)-e_g-\tfrac12\bar d-\bar\lambda\}
      \le a_g^\star,\\
U_r&:=\min\{1,\hat a_r+e_r\}\ge a_r^\star .
\end{align}
Monotonicity of clipping and the positive stabilizer then give
\[
Q^\star=\mathrm{clip}_{[0,1]}\!\left(\frac{a_g^\star}{a_r^\star+\varepsilon}\right)
\ge \mathrm{clip}_{[0,1]}\!\left(\frac{L_g}{U_r+\varepsilon}\right).
\]
The same two accuracy limits imply $a_g^\star-a_r^\star\ge L_g-U_r$.
The reported generated-only task score $\rho_2^{(i)}$ is evaluated on
$D_{\text{test}}$. The generated hold-out enters the domain-adaptation bound.

\subsection{Utility Bound}
\label{app:theory-utility}
\label{appendix:utility-bound}

Let $h_r$, $h_{rr}$, and $h_{rg}$ be trained on baseline real data, baseline
plus additional real data, and baseline plus generated data, respectively.
For the accuracy version of Utility, define
\begin{align}
\Delta^\star_{\mathrm{real}}&=R_P(h_r)-R_P(h_{rr}),\\
\Delta^\star_{\mathrm{generated}}&=R_P(h_r)-R_P(h_{rg}).
\label{eq:pop-gains}
\end{align}
The empirical gains $\hat\Delta_{\mathrm{real}}$ and
$\hat\Delta_{\mathrm{generated}}$ use the same real test set.

\paragraph{Comparing the gains.}
By \eqref{eq:DA-inst}, the population gains obey
\begin{equation}
\Delta^\star_{\mathrm{generated}}\ge \Delta^\star_{\mathrm{real}}
-\gamma^\star-\mathrm{disc}_{\mathcal H\Delta\mathcal H}(P,Q)-2\lambda(P,Q),
\qquad
\gamma^\star:=|R_Q(h_{rg})-R_Q(h_{rr})|.
\label{eq:pop-gap-final}
\end{equation}
To see this, subtract the two gains so that $R_P(h_r)$ cancels. Apply
\eqref{eq:DA-inst} to $h_{rr}$ and $h_{rg}$ and bound the difference of their
generated-domain risks below by $-\gamma^\star$.

\paragraph{Estimating the comparison terms.}
If the three predictors are fixed independently of the $n_t$ independent
real test observations drawn from $P$, Hoeffding's inequality~\citep{hoeffding1963probability}
and a union bound give
\begin{equation}
|\hat R_P(h)-R_P(h)|\le e_t,\quad h\in\{h_r,h_{rr},h_{rg}\},
\qquad
e_t=\sqrt{\frac{\log(6/\delta_t)}{2n_t}} .
\label{eq:gen}
\end{equation}
For the generated-domain gap, use a generated hold-out independent of the
two augmented predictors. If both generated-risk estimates have error at
most $e_g$, then
$\bar\gamma=|\hat R_Q(h_{rg})-\hat R_Q(h_{rr})|+2e_g$
upper-bounds $\gamma^\star$. The deterministic value $\bar\gamma=1$ is
available without this hold-out. Define $B=\bar\gamma+\bar d+2\bar\lambda$.
When all these bounds hold jointly, \eqref{eq:pop-gap-final} gives
\begin{equation}
\hat\Delta_{\mathrm{generated}}\ge\hat\Delta_{\mathrm{real}}-B-2e_t .
\label{eq:emp-gap}
\end{equation}
Only two real-domain risks enter this last sampling error because the
baseline is shared.

\paragraph{Normalizing the gains.}
For the empirical ratio, divide \eqref{eq:emp-gap} by
$\hat\Delta_{\mathrm{real}}$ when it exceeds $\tau$, then apply the clipping
in the score definition. The population score is
$U^\star=\mathrm{clip}_{[0,1]}(\Delta^\star_{\mathrm{generated}}/
\Delta^\star_{\mathrm{real}})$ when $\Delta^\star_{\mathrm{real}}>\tau$. Let
$L_{\mathrm{real}}=\hat\Delta_{\mathrm{real}}-2e_t$.
If $L_{\mathrm{real}}>\tau$, then
$\Delta^\star_{\mathrm{real}}\ge L_{\mathrm{real}}>\tau$ and
\begin{equation}
U^\star\ge
\mathrm{clip}_{[0,1]}\!\left(1-\frac{B}{L_{\mathrm{real}}}\right).
\label{eq:util-final}
\end{equation}
If the lower limit for the real-data gain does not exceed $\tau$, the
argument does not establish a lower bound for the population ratio. The reported
empirical score follows the observed-gain rule in
Section~\ref{sec:utility}.

\subsection{Indistinguishability Bound}
\label{app:theory-indistinguishability}

Let $a$ be the balanced accuracy of the fitted discriminator and $\hat a$
its observed value. Pool sizes of $N_{\mathrm{disc}}/2$ per source and
independence across all held-out observations make $\hat a$ the mean of
$N_{\mathrm{disc}}$ independent bounded correctness indicators.
Hoeffding's inequality gives
\[
|\hat a-a|\le e_I,\qquad
e_I=\sqrt{\frac{\log(2/\delta_I)}{2N_{\mathrm{disc}}}}
\]
with probability at least $1-\delta_I$. 
A different valid uniform radius can be obtained from a justified capacity
bound and sampling design \citep{mohri2018foundations}.
Because $g(a)=1-2|a-0.5|$ is $2$-Lipschitz, the lower limit for
Indistinguishability is $\max\{0,g(\hat a)-2e_I\}$.

\subsection{Similarity Bound}
\label{app:theory-similarity}
\label{appendix:similarity}

For each class, the probe selects the $k$ nearest non-self neighbors using
cosine similarity in the chosen representation. It computes the entropy from the observed
proportion of generated samples $\hat p_{cj}$ among the neighbors and
reports the proportion of generated samples in the pooled class.

\subsubsection{Conditional neighborhood sampling model}
\label{app:similarity-bound}

For a class $c$ and query $j$, assign an indicator equal to one for a
generated sample and zero for a real sample. Suppose the $k$ observed
indicators have the law of a simple random sample without replacement from
a specified finite reference population, conditional on the quantities used
to set the neighborhood. Let $p_{cj}^\star$ be the proportion of generated
samples in that population. This sampling assumption is not implied by
fixing the representation or selecting geometric nearest neighbors alone. Under this model, the two-sided
consequence of \citet[Corollary~1.1]{serfling1974}, with the finite-population
correction dropped for a looser radius, gives
\[
\Pr\bigl(|\hat p_{cj}-p_{cj}^\star|>r_c\bigr)\le\delta_c/N_c,
\qquad
r_c=\sqrt{\frac{\log(2N_c/\delta_c)}{2k}}.
\]
A union bound gives simultaneous intervals
$[\ell_{cj},u_{cj}]=[\max(0,\hat p_{cj}-r_c),\min(1,\hat p_{cj}+r_c)]$
for all queries in class $c$, without requiring independence across queries.

\paragraph{From proportions of generated samples to entropy.}
The minimum of the concave function $H_2$ over each interval is the smaller
of its two endpoint values. Its maximum is at the point in the interval
closest to $1/2$. These rules apply at proportions of zero and one. Averaging
the minima and maxima over the queries gives an interval for
$S_c^\star=N_c^{-1}\sum_j H_2(p_{cj}^\star)$ under the conditional model.
If $\sum_c\delta_c\le\delta_S$, applying the geometric mean to the classwise
lower and upper limits yields a simultaneous interval for
$S^\star=\operatorname{gmean}_{c\in\mathcal C}S_c^\star$ on the chosen
query pools. The coverage statement holds under the conditional sampling
model. For geometric neighborhoods where that model has not been
established, the corresponding screen is a diagnostic criterion rather
than a confidence bound.

\subsection{Composite Summary Bound}
\label{app:theory-exchangeability}
\label{appendix:exchangeability_metric}

The \emph{Exchangeability} field summarizes a complete profile. If
$L_Q,L_U,L_I,L_S$ are simultaneous lower limits for their population
scores, monotonicity gives
\[
\mathcal E_{\min}^\star\ge
\min\!\left\{\frac{L_Q+L_U}{2},\frac{L_I+L_S}{2}\right\}.
\]
The confidence level follows from the joint event for the four limits.
The label Exchangeability refers to this numerical summary.

%% file: Appendix.tex
E
min

=
min

{
Q
+
U
2
,
I
+
S
2
}
.

E
p
r
o
d
u
c
t
=
(
Q
+
U
2
)
(
I
+
S
2
)
.

E
min

=
min

{
Q
+
U
2
,
I
+
S
2
}
.

E
min

% =============================
% Appendix: Realigned / Cleaned
% =============================
% This appendix fragment is meant to be included into an existing LaTeX paper.
% It removes duplicate tables, fixes typos, aligns math, and standardizes float placement.
% 

% Ensure your bibliography includes the cited entries.

\section{Model Ranking Methodology}
\label{appendix:ranking}

\subsection{Ranking of Generative Models}
The ranking procedure uses the four diagnostic metrics: \emph{Quality},
\emph{Utility}, \emph{Indistinguishability}, and \emph{Similarity}.
For each dataset and generator, repeated evaluation resamples provide
an empirical distribution of each metric, summarized by its median and
interquartile range (IQR). Generators with complete profiles are grouped
by the number of failed screens on the four metrics, with preference given
to those passing more screens. Within each group, Monte Carlo pairwise
dominance incorporates the variability of the metric estimates into
the ranking.

\paragraph{Composite summary and incomplete profiles.}
We also report an \emph{Exchangeability} summary score (Sec.~\ref{appendix:exchangeability_metric}), but it is \emph{not} used in the pairwise domination test below to avoid double-counting, since it is derived from the four base metrics.

The four-metric ordering is computed only for profiles with an available normalized Utility value. When the matched real-data gain does not exceed $\tau$, Utility is unavailable, Exchangeability is withheld, and the profile is reported without assigning a four-metric domination rank. Utility Gain may still be reported as a separate augmentation quantity, but it is not substituted for normalized Utility in this ordering.

Let there be $G$ generators and $K$ base metrics (here $K=4$). Each generator $M_i$ has a performance vector
\begin{equation}
    \boldsymbol{\mu}_i = (\mu_{i1}, \mu_{i2}, \ldots, \mu_{iK}),
\end{equation}
with standard deviations
\begin{equation}
    \boldsymbol{\sigma}_i = (\sigma_{i1}, \sigma_{i2}, \ldots, \sigma_{iK}),
\end{equation}
describing variability across evaluation resamples for the fixed generator.
\paragraph{Quantile-to-moment inputs.}
The quantities $(\mu_{ik},\sigma_{ik})$ used below are the robust
pseudo-mean and pseudo-standard deviation derived from the empirical
quartiles $(Q_1,M,Q_3)$ as described in Appendix~\ref{app:quantile-to-moment}.
We use these $(\mu_{ik},\sigma_{ik})$ only to parameterise the Monte Carlo
sampling; the primary reported summaries for each model--metric pair remain
the median and IQR.
At each Monte Carlo iteration, the procedure draws independently for each
generator and metric and clips the results to the score range.
\begin{equation}
    \widetilde Z^{(t)}_{ik} \sim \mathcal{N}(\mu_{ik}, \sigma_{ik}^2),
    \qquad Z^{(t)}_{ik}=\min\{1,\max\{0,\widetilde Z^{(t)}_{ik}\}\}.
\end{equation}
A zero scale gives a point mass at $\mu_{ik}$. The independent sampling
model uses marginal variability across resamples rather than their joint
dependence structure. Dominance frequencies describe comparisons under
this sampling model; they are not confidence probabilities for a population
ordering.
Model $M_i$ \emph{dominates} $M_j$ at iteration $t$ iff $Z^{(t)}_{ik} \ge Z^{(t)}_{jk}$ for all $k$ and $Z^{(t)}_{ik} > Z^{(t)}_{jk}$ for at least one $k$. We accumulate domination counts $D_{ij}$.

\begin{algorithm}[!htbp]
\caption{Uncertainty-aware ranking of generative models}
\label{alg:ranking}
\begin{algorithmic}[1]
\REQUIRE Models $\mathcal M=\{M_1,\ldots,M_G\}$; failed-screen counts $f_i$; means $\boldsymbol{\mu}_i\in[0,1]^K$; stds $\boldsymbol{\sigma}_i\ge 0$; iterations $T$
\ENSURE Ranked list of models (using the four base metrics only)
\STATE Group models by $f_i$ in ascending order; count failed screens on the four base metrics only.
\FOR{each failure-count group $\mathcal G$}
  \STATE Initialize domination count matrix $D \gets \mathbf{0}_{|\mathcal G|\times |\mathcal G|}$
  \FOR{$t = 1$ to $T$}
    \FOR{each $M_i\in\mathcal G$}
      \FOR{$k = 1$ to $K$}
        \STATE Draw $\widetilde Z^{(t)}_{ik} \sim \mathcal{N}(\mu_{ik}, \sigma_{ik}^2)$; set $Z^{(t)}_{ik} \gets \min\{1,\max\{0,\widetilde Z^{(t)}_{ik}\}\}$
      \ENDFOR
    \ENDFOR
    \FOR{each ordered pair $(M_i,M_j)\in\mathcal G$, $i\neq j$}
      \IF{$Z^{(t)}_{ik} \ge Z^{(t)}_{jk}$ $\forall k$ and $\exists k: Z^{(t)}_{ik} > Z^{(t)}_{jk}$}
        \STATE $D_{ij} \gets D_{ij} + 1$
      \ENDIF
    \ENDFOR
  \ENDFOR
  \STATE $R_i \gets \sum_{j:M_j\in\mathcal G} D_{ij}$; sort $\mathcal G$ by descending $R_i$
\ENDFOR
\STATE Return the ranking together with the individual metric profiles and screening outcomes.
\end{algorithmic}
\end{algorithm}

\FloatBarrier

\section{Default Task Classifiers}
\label{appendix:default-classifiers}

\method uses a task classifier to compute Quality and Utility.
Indistinguishability uses a separate real-versus-generated discriminator, and
Similarity uses the binary entropy of the proportion of generated samples
in class-conditional nearest-neighbor neighborhoods.The synthetic experiments additionally report five intermediate performance quantities, denoted
$\rho_1,\ldots,\rho_5$: real-only, generated-only, real-plus-reference,
real-plus-generated, and real-versus-generated discriminator performance.
To avoid confounding generator differences with classifier changes, we
use a fixed task-classifier configuration within each reported experimental
setting, except where explicitly overridden. The task classifier is an
evaluation instrument: the comparison concerns its performance under
different training-data configurations, rather than optimizing its absolute
performance. The scores remain conditional on the selected classifier and
its ability to learn the task.
Experiments use random seed $42$ unless otherwise stated.

\paragraph{Real-data experiments (molecules and images).}
The reported molecular and image benchmarks use the following fixed
task-classifier configuration:

\begin{itemize}
    \item \textbf{Classifier:} ExtraTreesClassifier (scikit-learn)~\citep{geurts2006extratrees}
    \item \textbf{Number of trees:} 300
    \item \textbf{Parallelization:} \texttt{n\_jobs = -1}
    \item \textbf{Criterion:} Gini impurity
    \item \textbf{Bootstrap:} False
    \item \textbf{Max features:} scikit-learn default (all features, \texttt{max\_features=1.0})
    \item \textbf{Scoring:} \textbf{F1 score} on the held-out test set
\end{itemize}

This task-classifier configuration is fixed across the Quality and Utility
comparisons in the reported real-data experiments. The discriminator and
nearest-neighbor settings used for Indistinguishability and Similarity are
reported separately.

\paragraph{Controlled perturbation experiments.}
For the synthetic perturbation experiments (Sec.~\ref{sec:synthetic-quantile}),
the \emph{default} task classifier is:

\begin{itemize}
    \item \textbf{Classifier:} $k$-Nearest Neighbors (KNN)
    \item \textbf{Distance metric:} Euclidean
    \item \textbf{Number of neighbors:} $k=5$
    \item \textbf{Scoring:} \textbf{F1 score}
\end{itemize}

\section{Perturbation Techniques Applied to the Synthetically Generated Dataset}
\label{appendix:perturbation_descriptions}
We use controlled perturbations to investigate the response of the metrics
to label noise, class imbalance, copying, mode collapse, and feature
corruption. We compare the direction of each observed response with the
pre-specified hypotheses in Section~\ref{sec:synthetic-quantile}.

The experiments start with \( D_{\text{generated}} \approx D_{\text{train}} \)
and use a perturbation degree \( t \in [0, 1] \) to control severity.
We orient Spearman correlations so that positive values indicate
degradation as severity increases
(Appendix~\ref{app:standard-metric-comparison}). We compare the observed
directions with the expectations in Section~\ref{sec:synthetic-quantile}.

We applied the following perturbation techniques:

\begin{itemize}
    \item \textbf{Mode Collapse:} For each class, we select one representative generated
    instance and replace a controlled fraction of same-class instances with exact copies
    of that representative. This reduces within-class diversity while preserving class counts.
    \item \textbf{Class Imbalance:} Applied independently to every fixed-size split by resampling class labels to a specified positive-class ratio; the split sizes are preserved, with replacement used when a class is undersupplied.
    \item \textbf{Label Noise:} Added by randomly flipping a fraction of the target labels in the dataset.
    \item \textbf{Exact Copy Noise:} Introduced by replacing a subset of the \( D_{\text{generated}} \) instances with copies of samples drawn from \( D_{\text{train}} \).
    \item \textbf{Noise Copies:} A subset of \( D_{\text{generated}} \) is replaced with near-duplicates of real training samples from \( D_{\text{train}} \), each perturbed by zero-mean Gaussian noise with standard deviation \( \sigma = 0.05 \).
    \item \textbf{Gaussian Noise:} In the matched comparison, severity $s\in[0,1]$ selects $\operatorname{round}(s n)$ generated rows and adds independent Gaussian noise $\mathcal{N}(0,(0.6s)^2 I)$ to their features; labels and the real/test splits remain fixed. The separate feature-noise stress test additionally applies additive and multiplicative noise, feature dropout, sign flips, feature shuffling, extreme-value overwrites, and an optional nonlinear transform.
\end{itemize}
\section{Matched Comparison with Standard Metrics}
\label{app:standard-metric-comparison}

At each perturbation severity, GUIDE and the external measures use the same real reference set, generated samples, feature representation, and sample counts. The primary external measures are linear- and RBF-kernel MMD~\citep{gretton2012kernel}, precision, recall, density, and coverage~\citep{sajjadi2018assessing,kynkaanniemi2019improved,naeem2020reliable}, and exact RBF Vendi~\citep{friedman2023vendi}. The external measures and the real-versus-generated discriminator use features. Quality, Utility, and class-conditional Similarity also use task labels. GUIDE's Indistinguishability provides the classifier two-sample comparison.

Here Vendi denotes the exact order-one Vendi Score with an RBF kernel. For a trace-normalized kernel matrix with eigenvalues $\lambda_j$, order-two Vendi equals $(\sum_j\lambda_j^2)^{-1}$, the effective-count form of R\'enyi kernel entropy (RKE). The corresponding entropy is the natural logarithm of that value~\citep{pasarkar2024cousins,jalali2023rke}. The image experiments also report FID, relative RKE, and Fourier-based approximations to the Vendi and order-two effective counts~\citep{jalali2023rke,ospanov2024fkea}.

We report the four GUIDE scores, Exchangeability, and Utility Gain, $a_{rg}-a_r$ (
$\rho_4-\rho_1$). At or below the real-data gain cutoff $\tau=10^{-3}$, we report Utility Gain and leave normalized Utility and Exchangeability unavailable (Section~\ref{sec:utility}). The raw discriminator output $\rho_5$ is balanced accuracy. Indistinguishability, $1-2|\rho_5-0.5|$, equals one at chance accuracy and zero for perfect separation in either label orientation.

\begin{figure}[!htbp]
  \centering
  \includegraphics[width=0.75\linewidth]{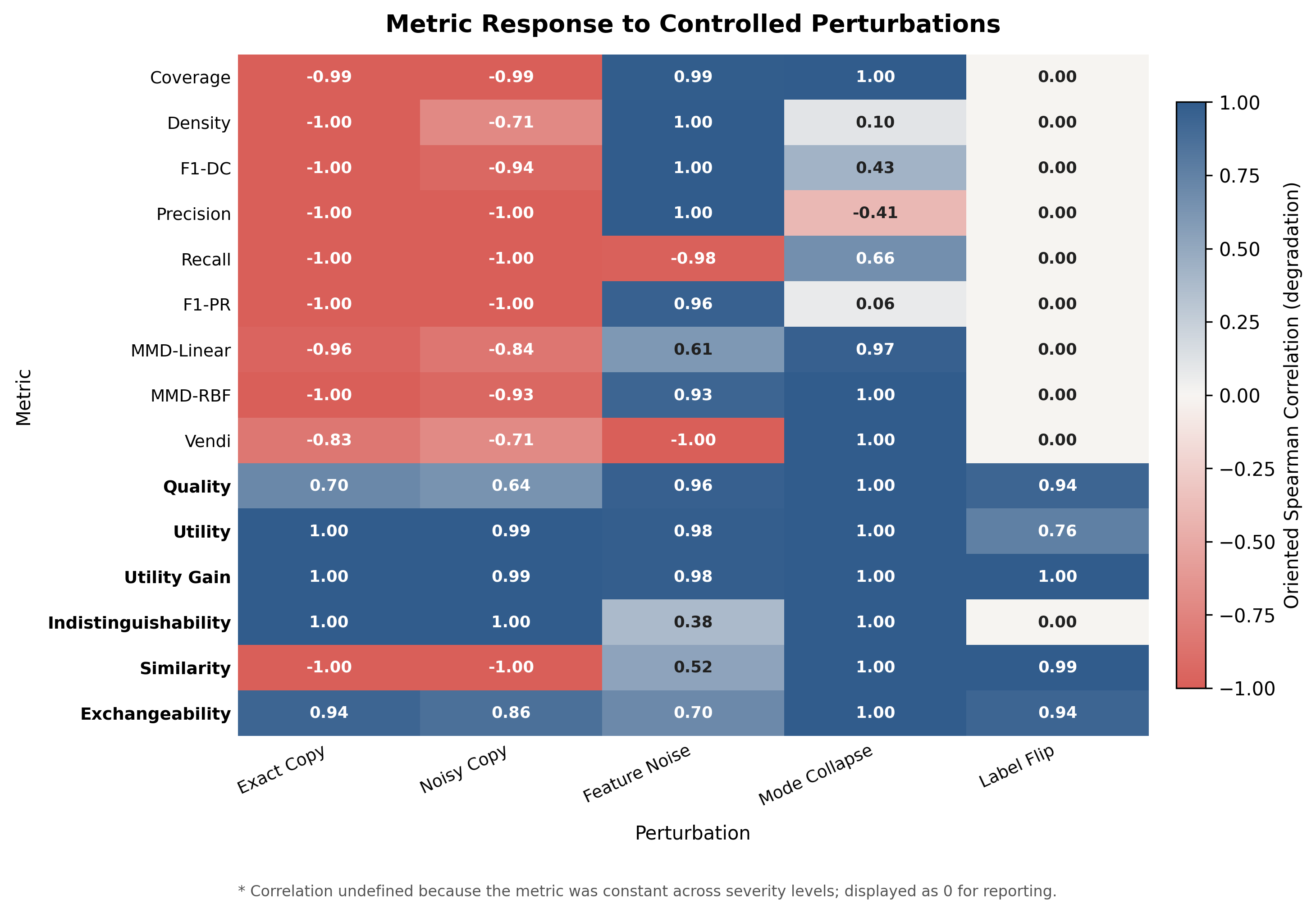}
  \caption{Spearman correlations between metric degradation and perturbation severity. Positive values indicate that a metric moves in the direction defined as degradation. The plot displays undefined correlations for constant scores as $0.00$.}
  \label{fig:matched-metric-comparison}
\end{figure}

Figure~\ref{fig:matched-metric-comparison} reports Spearman correlations with
severity, using the negative of each higher-is-better score and the original
value of each lower-is-better score. We compare the sign of each correlation
with the expected response in Section~\ref{sec:synthetic-quantile}. The
magnitude describes the consistency of the monotone response. Assessing the
size of a score change requires the score values across severities.

\paragraph{Copying.}
Utility and Utility Gain fall as copies are added, with oriented
correlations of $1.00$ for exact copying and $0.99$ for noisy copying.
Quality also tends to decline, with correlations of $0.70$ and $0.64$.
Both MMD variants, precision, recall, density, coverage, and Vendi move
in their favorable directions. Similarity increases in both copying
conditions, giving a correlation of $-1.00$ in each. The heatmap reports
$+1.00$ for Indistinguishability in both conditions, corresponding to a
decline in that score under the stated orientation. The sensitivity sweep
gives negative Indistinguishability
correlations for both copy types (Table~\ref{tab:spearman-compact}).
Both analyses show more within-class mixing as the predictive gain falls.

\paragraph{Feature noise and collapse.} With Gaussian feature noise, the degradation correlations are $0.96$ for Quality, $0.98$ for Utility and Utility Gain, and $0.52$ for Similarity. For the class-conditional collapse construction, all displayed GUIDE correlations round to $1.00$. This describes the response to concentrating each class around a single representative in this experiment.

\paragraph{Label flipping.} Under label flipping, the features stay fixed. The correlations are $0.94$ for Quality, $0.76$ for Utility, $1.00$ for Utility Gain, $0.99$ for class-conditional Similarity, and $0.94$ for Exchangeability. The feature-only scores and global Indistinguishability are constant in this sweep. The figure displays their undefined correlations as $0.00$.

\FloatBarrier

\section{Sensitivity Analysis: Classifiers, Dataset Sizes, and Scoring Rules}
\label{appendix:Experiments}

The primary matched perturbation analysis uses the fixed synthetic configuration
described in the main text. The sensitivity analysis uses 40 features
and varies dataset sizes, classifier families, and scoring rules. The class-imbalance analysis runs sizes
$500$, $1000$, $3000$, $5000$, and $10000$; the other five analyses use
$5000$ examples per split. This evaluation grid uses ExtraTrees (300 trees),
RandomForest (300 trees), GradientBoosting (100 estimators), AdaBoost (50
estimators), LogisticRegression, DecisionTree, SVC, KNN ($k=5$), and
GaussianNB, together with the listed F1, balanced-accuracy, precision, recall,
ROC-AUC, Jaccard, zero-one, log-loss, average-precision, Hamming, and hinge
scorers where enabled. This analysis examines the sensitivity of the metrics
to evaluation choices, complementing the matched comparison, which fixes
the task classifier at KNN with $k=5$ and uses binary F1.

The class-imbalance sweep uses positive-class ratios
$\{0,0.05,0.10,\ldots,0.40,0.50,0.60,\ldots,0.90,0.95,1\}$.
The other perturbation analyses in the sensitivity sweep use a balanced ratio of $0.5$. Their
original severity grid is $\{0,0.1,\ldots,1\}$; the matched analysis
uses twelve evenly spaced levels, including both endpoints. Table~\ref{tab:spearman-compact} summarizes how the GUIDE scores respond across the one-factor variations.

\begin{longtable}{|c|c|c|c|}
\hline
\textbf{Dataset Size} & \textbf{Classifier} & \textbf{Scorer} & \textbf{Positive Class Ratio} \\
\hline
\endfirsthead
\hline
\textbf{Dataset Size} & \textbf{Classifier} & \textbf{Scorer} & \textbf{Positive Class Ratio} \\
\hline
\endhead
\hline
\endfoot
100  & ExtraTreesClassifier           & F1 Score            & 0.40 \\
500  & RandomForestClassifier         & Balanced Accuracy   & 0.50 \\
1000 & GradientBoostingClassifier     & Precision           & 0.20 \\
2500  & AdaBoostClassifier             & Recall              & 0.60 \\
5000  & LogisticRegression             & ROC-AUC Score       & 0.35 \\
10000 & DecisionTreeClassifier         & Jaccard Score       & 0.25 \\
 & SVC                            &    Average Precision    & 0.90 \\
 & KNeighborsClassifier           &  Precision           & 0.30 \\
 & GaussianNB                     &    & 0.80 \\
 & ExtraTreesClassifier           &         & 0.95 \\
 & RandomForestClassifier         &         & 0.15 \\
 & GradientBoostingClassifier     &           & 0.10 \\
\hline
\end{longtable}

\subsection{Sensitivity-Analysis Settings}
\begin{small}
\begin{longtable}{|p{3.2cm}|p{2.5cm}|p{7.2cm}|}
\hline
\textbf{Parameter} & \textbf{Value} & \textbf{Description} \\
\hline
\endfirsthead
\hline
\textbf{Parameter} & \textbf{Value} & \textbf{Description} \\
\hline
\endhead
\hline
\endfoot
\textbf{Resampling iterations} & 100 (30 for class-imbalance sizes $\geq3000$) & Number of resampling iterations used in the sensitivity analysis. \\
\textbf{Instance resampling} & True & Whether instances are resampled during evaluation. \\
\textbf{Sampling with replacement} & False & Whether resampling uses replacement. \\
\textbf{Training fraction} & 0.8 & Fraction of each resample used to fit the task estimator. \\
\textbf{Task estimator} & KNeighbors\allowbreak{}Classifier & Estimator used to measure downstream task performance. \\
\textbf{Task score} & F1, AUC, etc. & Scores used to evaluate downstream predictive performance. \\
Reports the five $\rho$ quantities and the four GUIDE summaries for inspection. \\
 
\hline
\end{longtable}
\end{small}

\FloatBarrier

\section{Vectorizers Used in the Graph Data Experiments}
\label{appendix:Vectorizers}

In this study, we utilized several vectorizers for our  experiments. Below is a brief overview of each method used:

- GIN (Xu et al., 2019) \cite{xu2019powerful}: The Graph Isomorphism Network (GIN) is a  graph neural network  based on the idea of learning a graph's structural properties through a series of message-passing layers. GIN has demonstrated strong performance on graph classification tasks, especially when combined with sufficient training data and appropriate network architectures.

- GraphCL (You et al., 2020) \cite{DBLP:conf/nips/YouCSCWS20}: GraphCL is a contrastive learning method for graph data, which leverages a self-supervised learning framework that maximizes the agreement between augmented views of a graph while minimizing the distance between distinct views. This technique helps capture better graph representations, even without labeled data, making it suitable for semi-supervised learning settings.

- InfoGraph (Sun et al., 2020) \cite{sun2019infograph}: InfoGraph is a method for unsupervised and semi-supervised graph learning that focuses on learning graph-level representations. It uses mutual information maximization to capture global graph structures, enabling the model to generate meaningful and informative embeddings. 

- GraphVectorizer (Custom Implementation): The GraphVectorizer processes graph structures derived from SMILES strings. It extracts few graph features, including degree histograms, clustering coefficients, and node labels. These features are then encoded into fixed-length vectors, capturing graph topology and node-level information. The vectorizer includes padding to standardize feature lengths and applies label encoding to handle node labels.

- NSPDK~\citep{costa2010fast}: The Fast Neighborhood Subgraph Pairwise Distance Kernel (NSPDK) computes the similarity between two graphs based on the pairwise distances between their corresponding subgraphs. This kernel is designed to efficiently capture structural information by evaluating local neighborhoods and pairwise distances.

- HashedAPVectorizer \cite{rdkit2011}: Computes Atom Pair fingerprints by capturing relationships between atom pairs in a molecule, resulting in a fixed-length hashed vector. This vectorizer is used for molecular similarity tasks.

- HashedMorganVectorizer \cite{rdkit2011}: Generates Morgan fingerprints, which are circular and capture local atom environments. It is widely used in molecular similarity and classification tasks.

- HashedRDKVectorizer \cite{rdkit2011}: Computes RDKit Fingerprints, which are based on atom connectivity and properties. This vectorizer is suitable for general molecular similarity and topological feature representation.

- HashedTorsionVectorizer \cite{rdkit2011}: Generates a hashed topological torsion fingerprint from paths of bonded atoms in the molecular graph, with four atoms per path by default.

\section{Datasets}
\label{appendix:Datasets}

\subsection{Molecular Datasets} 
\label{appendix:molecular_datasets}

We conduct experiments using several datasets from the Therapeutic Data Commons \cite{huang2021tdc}, which are designed for single prediction tasks. By default, the datasets are split into training, testing, and validation sets with proportions of 70\%, 20\%, and 10\%, respectively, using random splitting. The raw data is initially provided as SMILES strings \cite{weininger1988smiles}. For this study, we apply multiple vectorization techniques, including both \textit{molecular} and \textit{graph-based} vectorizers.
The molecular vectorizers process the SMILES strings and generates molecular features, which are then used by our metric. Additionally, we employ the graph-based vectorizers to convert SMILES into graph representations, where nodes correspond to atoms and edges correspond to bonds in the molecule.
Each molecular graph is represented by \(G=(V,E,X,A)\), with \(N=|V|\) atoms and an edge list \(E\subseteq V\times V\) recording the bonds. The PyTorch Geometric encoding stores \(X\in\mathbb{Z}^{N\times 9}\) for atom features and \(A\in\mathbb{Z}^{|E|\times 3}\) for edge features, with one edge entry in each direction per bond. The atom features encode atomic number, chirality, degree, formal charge, hydrogen count, radical electron count, hybridization, aromaticity, and ring membership. The edge features encode bond type, stereochemistry, and conjugation~\citep{pyg_smiles_encoder}.

\begin{itemize} 
    \item \textbf{AMES Mutagenicity}: Consists of 7,255 binary drugs labeled as mutagenic or non-mutagenic with $3 \leq |V| \leq 55$. 
    \item \textbf{BBB Martins}: 1,975 drugs classified into those that can penetrate the blood-brain barrier (BBB) or not with $2 < |V| < 132$. 
    \item \textbf{CYP1A2 Veith}: 12,579 compounds labeled for CYP1A2 enzyme inhibition, with $7 \leq |V| \leq 123$.
    \item \textbf{CYP2C19 Veith}: 12,665 compounds labeled for CYP2C19 enzyme inhibition, with $7 \leq |V| \leq 114$.
    \item \textbf{Herg Karim}: Consists of 13,445 molecular structures labeled as hERG ($<10\,\mu M$) and non-hERG ($\geq 10\,\mu M$) blockers with $6 < |V| < 58$. 
\end{itemize}
Table~\ref{tab:molecular-splits-main} reports the split sizes used for the real-generator evaluation.

\begin{table}[!htb]
\centering
\fontsize{9}{10}\selectfont
\setlength{\tabcolsep}{2.5pt}

\caption{Molecular split sizes used for real-generator evaluation.
$\DtrainOne$ is used to train the generator, $\DtrainTwo$ provides the
real augmentation reference, and $\Dtest$ is held out for evaluation.}
\label{tab:molecular-splits-main}

\begin{tabular}{@{}lrrrr@{}}
\toprule
\textbf{Dataset}
& \textbf{$D_{\text{gen}}$}
& \textbf{$D_{\text{train1}}$}
& \textbf{$D_{\text{train2}}$}
& \textbf{$D_{\text{test}}$} \\
\midrule
AMES & 2340 & 2340 & 2340 & 1316 \\
BBB Martins & 318 & 318 & 318 & 206 \\
CYP1A2 Veith & 4128 & 4401 & 4128 & 2376 \\
CYP2C19 Veith & 3956 & 3956 & 3956 & 2326 \\
hERG Karim & 4700 & 4700 & 4700 & 2686 \\
\bottomrule
\end{tabular}

\end{table}

\begin{table}[htbp]
    \centering
    \small
    \caption{General statistics for molecular datasets used in our experiments. Node types are shown for positive and negative samples separately.}
    \begin{tabular}{|l|c|c|c|p{3.3cm}|c|}
        \hline
        \textbf{Dataset} & \textbf{Size} & \textbf{Positives} & \textbf{Negatives} & \textbf{Node Types (Pos, Neg)} & \textbf{Edge Types} \\
        \hline
        \texttt{ames} & 5094 & 2759 & 2335 & 10 (Pos), 10 (Neg) & 4 \\
        \texttt{bbb\_martins} & 1421 & 1096 & 325 & 12 (Pos), 11 (Neg) & 4 \\
        \texttt{cyp1a2\_veith} & 8805 & 4060 & 4745 & 17 (Pos), 25 (Neg) & 4 \\
        \texttt{cyp2c19\_veith} & 8866 & 4063 & 4803 & 17 (Pos), 25 (Neg) & 4 \\
        \texttt{herg\_karim} & 9412 & 4714 & 4698 & 10 (Pos), 15 (Neg) & 4 \\
        \hline
    \end{tabular}
\end{table}

\bigskip

\begin{table}[htbp]
    \centering
    \caption{Atomic Numbers for Positive and Negative Classes}
    \begin{tabular}{|l|}
        \hline
        \textbf{Atomic Numbers (Positive)} \\
        \hline
        $[1, 35, 6, 7, 8, 9, 15, 16, 17, 53]$ \\
        $[1, 5, 6, 7, 8, 9, 11, 15, 16, 17, 35, 53]$ \\
        $[1, 3, 6, 7, 8, 9, 11, 14, 15, 16, 17, 80, 78, 20, 24, 25, 26, 29, 30, 33, 34, 35, 50, 51, 53]$ \\
        $[1, 5, 6, 7, 8, 9, 11, 14, 15, 16, 17, 80, 19, 78, 26, 29, 35, 53]$ \\
        $[1, 35, 6, 7, 8, 9, 14, 16, 17, 53]$ \\
        $[1, 6, 7, 8, 9, 14, 15, 16, 17, 34, 35, 53]$ \\
        \hline
        \textbf{Atomic Numbers (Negative)} \\
        \hline
        $[1, 35, 6, 7, 8, 9, 15, 16, 17, 53]$ \\
        $[1, 35, 6, 7, 8, 9, 11, 15, 16, 17, 53]$ \\
        $[1, 5, 6, 7, 8, 9, 11, 78, 15, 16, 17, 14, 19, 20, 25, 26, 29, 30, 33, 34, 35, 44, 50, 51, 53]$ \\
        $[1, 5, 6, 7, 8, 9, 11, 78, 15, 16, 17, 14, 19, 20, 80, 25, 26, 29, 30, 33, 34, 35, 44, 50, 51, 53]$ \\
        $[1, 5, 6, 7, 8, 9, 11, 14, 15, 16, 17, 79, 34, 35, 53]$ \\
        $[1, 35, 5, 6, 7, 8, 9, 15, 16, 17]$ \\
        \hline
    \end{tabular}
\end{table}

\FloatBarrier

\subsection{SYNTHETIC Dataset}
The controlled dataset is generated with scikit-learn's \textbf{make\_classification} function using 40 informative, non-redundant features, four clusters per class, class separation 0.6, and no label noise. We first create four equal, stratified sets---training, reference, test, and generated---and fit the standardizer on the training set only. That fitted transformation is then applied unchanged to the reference, generated, and test sets. The generated split initially comes from the same controlled task distribution as the real splits. For copying, label noise, mode collapse, and feature noise, the perturbation is applied only to the generated split; the separate class-imbalance sweep rebalances all four fixed-size sets.

\subsection{Image Datasets}
We evaluate on three canonical benchmarks: \textbf{MNIST}~\citep{lecun1998gradient},
\textbf{Fashion\textendash MNIST}~\citep{xiao2017fashion}, and \textbf{CIFAR\textendash 10}~\citep{krizhevsky2009learning}.
All models share preprocessing and canonical stratified splits
(details in Appendix~\ref{app:image-experiments});
test images are never used for training or model selection.

\section{Generators Used}
\label{app:generators}

\subsection{Molecular(graph) data generation}
\label{app:molecular_generators}

In this study, we utilized several graph generation models for our experiments. Below is a brief overview of each method used, along with the corresponding citations:
\begin{table}[!htbp]
\centering
\caption{Overview of Generative Models Used in This Study}
\label{tab:generator_summary}
\renewcommand{\arraystretch}{1.2}
\setlength{\tabcolsep}{6pt}
\small
\begin{tabularx}{\linewidth}{@{}p{0.30\linewidth}p{0.18\linewidth}X@{}}
\toprule
\textbf{Generator} & \textbf{Model Class} & \textbf{Description} \\
\midrule
STGG~\cite{ahn2021spanning} & Autoregressive & Spanning-tree-based decoder with Transformer backbone \\
WGAN-GP~\cite{keras2021wgan} & GAN-based & Adversarial model using R-GCN for graph discrimination \\
JTNN~\cite{jin2018junction} & VAE-based & Junction-tree structured variational autoencoder \\
HierVAE~\cite{jin2020hiervae} & VAE-based & Hierarchical VAE with coarse-to-fine motif decoding \\
MoFlow~\cite{zang2020moflow} & Flow-based & Bond flow and a bond-conditioned atom flow \\
GDSS~\cite{jo2022score} & Diffusion-based & SDE-based generative diffusion over graphs \\
S4DD~\cite{ozcelik2024s4} & State-space & Chemical language model with recurrent sampling \\
SWINGNN~\cite{yan2023swingnn} & Diffusion-based & 2-WL guided diffusion with shifted-window attention \\
Neighborhood Swap (NS-1/2/3) & Perturbation-based & Iterative, non-parametric local rewiring for augmentation \\
\bottomrule
\end{tabularx}
\end{table}
\begin{itemize}
    \item \textbf{STGG (Spanning Tree Graph Generator)} \cite{ahn2021spanning}: Formulates the graph generation of a molecular graph as a sequence of tree-constructive operations applied through the composition of a spanning tree with the residual edges. STGG adopts a transformer architecture which generates the tree by using relative positional encodings and residual edges using an attention-based predictor.
    
    \item \textbf{WGAN-GP with R-GCN (Wasserstein GAN with Gradient Penalty and Relational Graph Convolutional Networks)} \cite{keras2021wgan}: Originally designed for the generation of small molecular graphs such as QM9, but adjusted to generate larger compounds. The generator network consists of two fully connected networks, and the discriminator implements non-linearly transformed neighborhood aggregations through the means of R-GCN.

    \item \textbf{JTNN (Junction Tree Variational Autoencoder for Molecular Graph Generation)} \cite{jin2018junction}: Generates graphs in two stages. In the first stage, a tree structured object is generated by exploiting the coarse-grained representations of the training molecular graphs. In the second stage, the nodes in the tree (which are essentially subgraphs) are assembled back into molecules.

    \item \textbf{HierVAE} \cite{jin2020hiervae}: Uses a hierarchical encoder-decoder architecture for graph generation. The encoder produces a multi-resolution representation for each input graph in a fine to coarse fashion, starting with the atoms and finishing with fully connected motifs. During the decoding process, the molecules are assembled back in a coarse to fine manner, where three consecutive predictions are made at each pass: new motif selection, which part of it attaches, and the points of contact with the current molecule.

    \item \textbf{MoFlow} \cite{zang2020moflow} uses a Glow-based flow for bonds and a flow for atoms conditioned on the bonds. Sampling inverts the two flows and applies a validity correction.

    \item \textbf{GDSS} \cite{jo2022score}: Proposes a continuous-time SDE system to model the diffusion process over nodes and edges simultaneously, where Gaussian noise is directly inserted into the adjacency matrix and node features. Sampling is done by solving the SDE used to describe the reverse-time diffusion process.

    \item \textbf{S4DD} \cite{ozcelik2024s4} is a chemical language model based on the S4 state-space architecture. It uses the convolutional formulation during training and the recurrent formulation to generate sequences. We evaluate three configurations:
\begin{itemize}
    \item \texttt{unfiltered\_1}: trained directly on the target dataset without pretraining or filtering.
    \item \texttt{unfiltered\_m}: pretrained on a large, diverse set of over 1 million molecules (e.g., toxic and non-toxic compounds from the Therapeutics Data Commons), then fine-tuned.
    \item \texttt{filtered\_m}: pretrained on a combined dataset (ZINC + QM9 + ChEMBL), filtered to include only common atom types, then fine-tuned.
    
\end{itemize}
    \item \textbf{SWINGNN} \cite{yan2023swingnn}: A non-invariant diffusion model that employs an edge-to-edge 2-WL message passing network and utilizes shifted window-based self-attention. They propose a 2nd order sampler with correction for generating large molecular graphs.

    \item \textbf{Neighborhood Swap Generator (NS)}: A non-parametric, perturbation-based generator that produces augmented graphs through iterative local rewiring. It begins by decomposing each input graph into small neighborhoods (e.g., 1--2-hop subgraphs) and then perturbs them via controlled rewiring and recombination. The generator operates in multiple sequential rounds, where each round applies perturbations not only to the original structure but also to neighborhoods altered in previous steps. This compounding process introduces increasingly diverse graph structures while preserving essential properties. We define three versions---NS-1, NS-2, and NS-3---corresponding to one, two, and three rounds of perturbation, respectively. NSG requires no training and is well-suited for data augmentation and robustness evaluation in graph learning tasks.
\end{itemize}

\subsubsection{Hyperparameters for molecular graph generators}
\label{sec:hyperparameters}

\begin{table}[!htbp]
\centering
\small
\caption{HierVAE Hyperparameters}
\begin{tabular}{ll|ll}
\toprule
Parameter & Value & Parameter & Value \\
\midrule
\texttt{rnn\_type} & LSTM & \texttt{depthG} & 15 \\
\texttt{hidden\_size} & 250 & \texttt{diterT} & 1 \\
\texttt{embed\_size} & 250 & \texttt{diterG} & 3 \\
\texttt{batch\_size} & 50 & \texttt{dropout} & 0 \\
\texttt{latent\_size} & 32 & \texttt{learning\_rate} & $1 \times 10^{-3}$ \\
\texttt{depthT} & 15 & \texttt{clip\_norm} & 5 \\
\texttt{step\_beta} & $1 \times 10^{-3}$ & \texttt{max\_beta} & 1 \\
\texttt{warmup} & 10000 & \texttt{kl\_anneal\_rate} & 0.9 \\
\texttt{epochs} & 25000 & -- & -- \\
\bottomrule
\end{tabular}
\end{table}
\FloatBarrier

\begin{table}[!htbp]
\centering
\small
\caption{JTNN Hyperparameters}
\begin{tabular}{ll|ll}
\toprule
Parameter & Value & Parameter & Value \\
\midrule
\texttt{epochs} & 5 & \texttt{beta} & 0 \\
\texttt{learning\_rate} & $1 \times 10^{-3}$ & \texttt{max\_beta} & 1 \\
\texttt{annealing\_rate} & 0.9 & \texttt{save\_iter} & 5000 \\
\texttt{clip\_norm} & 50 & \texttt{step\_beta} & 0.002 \\
\texttt{total\_step} & 0 & \texttt{annealing\_iterations} & 40000 \\
\texttt{kl\_anneal\_iter} & 2000 & \texttt{hidden\_size} & 56 \\
\texttt{latent\_size} & 40 & -- & -- \\
\bottomrule
\end{tabular}
\end{table}
\FloatBarrier

\begin{table}[!htbp]
\centering
\small
\caption{S4forDenovoDesign S4DD Hyperparameters}
\begin{tabular}{ll|ll}
\toprule
Parameter & Value & Parameter & Value \\
\midrule
\texttt{model\_dim} & 256 & \texttt{state\_dim} & 64 \\
\texttt{n\_layers} & 4 & \texttt{n\_ssm} & 1 \\
\texttt{dropout} & 0.25 & \texttt{vocab\_size} & 50 \\
\texttt{sequence\_length} & 120 & \texttt{n\_max\_epochs} & 400 \\
\texttt{learning\_rate} & $1 \times 10^{-3}$ & \texttt{batch\_size} & 500 \\
\texttt{device} & \texttt{"cuda"} & -- & -- \\
\bottomrule
\end{tabular}
\end{table}
\FloatBarrier

\begin{table}[!htbp]
\centering
\small
\caption{GDSS Model Hyperparameters}
\begin{tabular}{ll|ll}
\toprule
Parameter & Value & Parameter & Value \\
\midrule
\texttt{s\_theta} & 2 & \texttt{s\_phi} & 16 \\
\texttt{GCN\_layers} & 4 & \texttt{hidden\_dim} & 2 \\
\texttt{attention\_heads} & 8 & \texttt{initial\_channels} & 4 \\
\texttt{hidden\_channels} & 3 & \texttt{final\_channels} & 16 \\
\texttt{SDE\_X} & VP & \texttt{sampling\_steps\_X} & 1000 \\
$\beta_{\min}^{(X)}$ & 0.1 & $\beta_{\max}^{(X)}$ & 1 \\
\texttt{SDE\_A} & VE & \texttt{sampling\_steps\_A} & 1000 \\
$\beta_{\min}^{(A)}$ & 0.2 & $\beta_{\max}^{(A)}$ & 1 \\
\texttt{solver} & Rev + Langevin & -- & -- \\
\bottomrule
\end{tabular}
\end{table}
\FloatBarrier

\begin{table}[!htbp]
\centering
\small
\caption{GDSS Training Settings}
\begin{tabular}{lllllll}
\toprule
Setting & Optimizer & LR & Weight Decay & Batch & Epochs & EMA \\
\midrule
Mol Graphs & Adam & $5 \times 10^{-3}$ & $1 \times 10^{-4}$ & 56 & 1000 & 0.999 \\
\bottomrule
\end{tabular}
\end{table}
\FloatBarrier

\begin{table}[!htbp]
\centering
\small
\caption{STGG Hyperparameters}
\begin{tabular}{llllll}
\toprule
LR & Optimizer & Epochs & Layers & Embedding Size & Dropout \\
\midrule
0.0001 & Adam & 100 & 3 & 1024 & 0.1 \\
\bottomrule
\end{tabular}
\end{table}
\FloatBarrier

\begin{table}[!htbp]
\centering
\small
\caption{MoFlow Hyperparameters}
\resizebox{\textwidth}{!}{
\begin{tabular}{lllllll}
\toprule
Batch & LR & LR Decay & Epochs & \texttt{b\_n\_flow} & \texttt{b\_n\_block} & \texttt{b\_hidden\_ch} \\
\midrule
12 & 0.001 & 0.999995 & 200 & 10 & 1 & 128/128 \\
\bottomrule
\toprule
\texttt{b\_conv\_lu} & \texttt{a\_n\_flow} & \texttt{a\_n\_block} & \texttt{a\_hidden\_gnn} & \texttt{a\_hidden\_lin} & \texttt{learn\_dist} & \texttt{noise\_scale} \\
\midrule
1 & 27 & 1 & 64 & 128/64 & True & 0.6 \\
\bottomrule
\end{tabular}}
\end{table}
\FloatBarrier

\begin{table}[!htbp]
\centering
\small
\caption{Neighborhood Swap Generator Hyperparameters}
\resizebox{\textwidth}{!}{
\begin{tabular}{lccccccc}
\toprule
Version & \texttt{num\_iterations} & \texttt{min\_size} & \texttt{max\_size} & \texttt{nbits} & \texttt{size} & \texttt{max\_permutations} & \texttt{parallel} \\
\midrule
NS1 & 1 & 1 & 2 & 12 & 20 & 2 & True \\
NS2 & 2 & 1 & 2 & 12 & 20 & 2 & True \\
NS3 & 3 & 1 & 2 & 12 & 20 & 2 & True \\
\bottomrule
\end{tabular}}
\end{table}
\FloatBarrier

\begin{table}[!htbp]
\centering
\small
\caption{WGAN-GP Hyperparameters}
\begin{tabular}{ll}
\toprule
Parameter & Value \\
\midrule
Dense Units & 128, 256, 512 \\
Dropout Rate & 0.2 \\
Discriminator Steps & 1 \\
GConv Units & 128, 128, 128, 128 \\
Generator Dense Units & 512, 512 \\
Generator Dropout Rate & 0.2 \\
Generator Steps & 1 \\
Gradient Penalty Weight & 10 \\
Optimizer & Adam \\
Learning Rate & 0.001 \\
Epochs & 20 \\
Batch Size & 28 \\
\bottomrule
\end{tabular}
\end{table}
\FloatBarrier

\begin{table}[!htbp]
\centering
\small
\caption{SWINGNN Hyperparameters}
\begin{tabular}{llllll}
\toprule
Graph Type & MCMC & Model & Precond & Sigma Dist & Steps \\
\midrule
Molecular & edm & edm & edm & 256 & -1, 1, $x_0$ \\
\bottomrule
\toprule
Model Name & Feature Dims & Depths & Window Size & Patch Size & Sample Clip \\
\midrule
swin\_gnn & Molecular & 1 1 3 1 & 2 & 3 & -- \\
\bottomrule
\end{tabular}
\end{table}
\FloatBarrier

\subsection{Image data Generators}
\subsubsection{Generative Models and Training Setups}
\label{app:image-models}

We evaluate conditional VAEs~\citep{kingma2014auto,rezende2014stochastic,sohn2015conditional}, GAN variants (DCGAN~\citep{radford2016unsupervised}, WGAN-GP with projection
critic~\citep{gulrajani2017improved,miyato2018cgans}, StyleGAN2-lite adapted to $32{\times}32$~\citep{karras2020analyzing}), and conditional diffusion models
(UNet backbones~\citep{ronneberger2015u} and Transformer2D/DiT-style backbones~\citep{peebles2023di}) trained as
DDPMs~\citep{ho2020denoising} with classifier-free guidance.

\paragraph{Conditional VAEs.}
Shared training hyperparameters (per dataset) and compact architecture deltas are listed in
Tables~\ref{tab:vae-shared} and~\ref{tab:vae-arch}. All VAEs are class-conditional with label
embeddings concatenated at encoder/latent/decoder as appropriate.

\begin{table}[!htbp]
\centering
\scriptsize
\caption{VAE training setup shared across MLP/Conv/Res (per dataset).}
\label{tab:vae-shared}
\begin{adjustbox}{max width=\linewidth}
\begin{tabularx}{\linewidth}{@{}lcccccccccc@{}}
\toprule
Dataset & $z$ & Cond & Ep & Bs & LR & KW & $\beta$ & FB & Loss & Aug \\
\midrule
MNIST        & 32  & 32 & 80  & 128 & $1\times 10^{-3}$ & 20 & 1.0 & 0.0 & BCE & none \\
FashionMNIST & 32  & 32 & 100 & 128 & $1\times 10^{-3}$ & 25 & 1.0 & 0.0 & BCE & none \\
CIFAR-10     & 128 & 64 & 250 & 128 & $1\times 10^{-3}$ & 30 & 1.0 & 0.0 & MSE & flip\,+\,crop \\
\bottomrule
\end{tabularx}
\end{adjustbox}
\end{table}

\begin{table}[!htbp]
\centering
\scriptsize
\caption{VAE architecture details; decoders mirror encoders unless noted.}
\label{tab:vae-arch}
\begin{adjustbox}{max width=\linewidth}
\begin{tabularx}{\linewidth}{@{}l l X X@{}}
\toprule
Arch & Datasets & Encoder & Decoder / Notes \\
\midrule
MLP     & MNIST / FashionMNIST & FC: 512, 512; heads $\mu,\log\sigma^2$ ($z$ per Tbl.~\ref{tab:vae-shared}) & FC: 512, 512 $\to$ reshape to $1{\times}28{\times}28$ \\
MLP     & CIFAR-10   & FC: 1024, 1024 & FC: 1024, 1024 $\to$ $3{\times}32{\times}32$ \\
Conv    & MNIST / FashionMNIST & $3{\times}3$/s2: 32, 64, 128 & Deconv s2: 128, 64, 32 $\to C$ \\
Conv    & CIFAR-10   & $4{\times}4$/s2: 64, 128, 256, 256 & Deconv s2: 256, 256, 128, 64 $\to C$ \\
ResConv & MNIST / FashionMNIST & ResDown s2: 32, 64, 128 & ResUp mirror; final $3{\times}3$ to $C$ \\
ResConv & CIFAR-10   & ResDown s2: 64, 128, 256, 256 & ResUp mirror; final $3{\times}3$ to $C$ \\
\bottomrule
\end{tabularx}
\end{adjustbox}
\end{table}

\paragraph{WGAN-GP (projection critic).}
Tables~\ref{tab:wgangp-data}--\ref{tab:wgangp-arch} summarize preprocessing deviations, training
hyperparameters, and architectures.

\begin{table}[!htbp]
\centering
\scriptsize
\caption{\textbf{WGAN-GP} data preprocessing deviations relative to Table~\ref{tab:img-data-shared}.}
\label{tab:wgangp-data}
\begin{adjustbox}{max width=\linewidth}
\begin{tabularx}{\linewidth}{@{}lcccX@{}}
\toprule
Dataset & Train input & Export size & Aug & Notes \\
\midrule
MNIST        & $1{\times}32{\times}32$ ($28\!\to\!32$) & $28{\times}28$ & none & Center-crop $32\!\to\!28$ on export; scale $[-1,1]$. \\
FashionMNIST & $1{\times}32{\times}32$ ($28\!\to\!32$) & $28{\times}28$ & none & Center-crop $32\!\to\!28$ on export; scale $[-1,1]$. \\
CIFAR-10     & $3{\times}32{\times}32$                  & $32{\times}32$ & none & No cropping; scale $[-1,1]$. \\
\bottomrule
\end{tabularx}
\end{adjustbox}
\end{table}

\begin{table}[!htbp]
\centering
\scriptsize
\caption{\textbf{WGAN-GP} training hyperparameters (shared across datasets).}
\label{tab:wgangp-hparams}
\begin{adjustbox}{max width=\linewidth}
\begin{tabular}{@{}ccccccccccc@{}}
\toprule
Ep & Batch & LR & $z$ & Cond & $G$ base & $D$ base & $D$ feat & $n_{\text{critic}}$ & $\lambda_{\text{GP}}$ & Adam $(\beta_1,\beta_2)$ \\
\midrule
100 & 64 & $1{\times}10^{-4}$ & 128 & 128 & 256 & 64 & 256 & 5 & 10.0 & $(0.0,\,0.9)$ \\
\bottomrule
\end{tabular}
\end{adjustbox}
\end{table}

\begin{table}[!htbp]
\centering
\scriptsize
\caption{\textbf{WGAN-GP} model: DCGAN-style conditional generator and projection critic~\citep{miyato2018cgans}.}
\label{tab:wgangp-arch}
\begin{adjustbox}{max width=\linewidth}
\begin{tabularx}{\linewidth}{@{}lX@{}}
\toprule
Generator & $z{\in}\mathbb{R}^{128}$, label emb.\ $\in\mathbb{R}^{128}$; concat $\to$ FC to $4{\times}4{\times}256$; deconvs: $256{\to}128{\to}64{\to}32$ (stride 2, BN+ReLU); $3{\times}3$ conv $\to C$; $\tanh$ output in $[-1,1]$. \\[2pt]
Critic (proj) & Convs: $C{\to}64{\to}128{\to}256$ (stride 2, LeakyReLU 0.2) to $4{\times}4$; global sum pool $\to \mathbb{R}^{256}$; score $= w^\top h + \langle h,\,E[y]\rangle$ with class embedding $E\!\in\!\mathbb{R}^{K\times256}$. \\
Objective & WGAN-GP with projection discriminator; gradient penalty $\lambda_{\text{GP}}{=}10$; $n_{\text{critic}}{=}5$. \\
Exports & Per-class sampler: generates $[-1,1]$ then maps to $[0,1]$; MNIST/FashionMNIST center-cropped $32{\to}28$ on save; CIFAR-10 kept at $32{\times}32$. \\
\bottomrule
\end{tabularx}
\end{adjustbox}
\end{table}

\paragraph{DCGAN and StyleGAN2-lite.}
Preprocessing and training settings are summarized in Tables~\ref{tab:gans-data}--\ref{tab:gans-arch}.

\begin{table}[!htbp]
\centering
\scriptsize
\caption{Data preprocessing deviations for \textbf{DCGAN}~\citep{radford2016unsupervised} and \textbf{StyleGAN2-lite}~\citep{karras2020analyzing} relative to Table~\ref{tab:img-data-shared}.}
\label{tab:gans-data}
\begin{adjustbox}{max width=\linewidth}
\begin{tabularx}{\linewidth}{@{}lcccX@{}}
\toprule
Model & Dataset & Train input & Export size & Notes \\
\midrule
DCGAN & MNIST / FashionMNIST & $1{\times}28{\times}28$ & $28{\times}28$ & No aug; scale $[-1,1]$. \\
DCGAN & CIFAR-10             & $3{\times}32{\times}32$ & $32{\times}32$ & No aug; scale $[-1,1]$. \\
\midrule
StyleGAN2-lite & MNIST / FashionMNIST & $1{\times}32{\times}32$ ($28\!\to\!32$) & $32{\times}32$ & SG2-lite fixed to $32^2$; no export crop. \\
StyleGAN2-lite & CIFAR-10             & $3{\times}32{\times}32$                  & $32{\times}32$ & SG2-lite fixed to $32^2$. \\
\bottomrule
\end{tabularx}
\end{adjustbox}
\end{table}

\begin{table}[!htbp]
\centering
\scriptsize
\caption{Training hyperparameters for \textbf{DCGAN} and \textbf{StyleGAN2-lite}.}
\label{tab:gans-train}
\begin{adjustbox}{max width=\linewidth}
\begin{tabular}{@{}lcccccccccc@{}}
\toprule
Model & Dataset & Epochs & Batch & $z$ & Cond.\ dim & $\,\text{LR}_G$ & $\,\text{LR}_D$ & Adam $(\beta_1,\beta_2)$ & Reg. & EMA \\
\midrule
\multirow{3}{*}{DCGAN}
& MNIST        & 100 & 128 & 128 & 64 & $2{\times}10^{-4}$ & $2{\times}10^{-4}$ & $(0.5,\,0.999)$ & BCE (real=0.9) & $0.999$ \\
& FashionMNIST & 120 & 128 & 128 & 64 & $2{\times}10^{-4}$ & $2{\times}10^{-4}$ & $(0.5,\,0.999)$ & BCE (real=0.9) & $0.999$ \\
& CIFAR-10     & 200 & 128 & 128 & 64 & $2{\times}10^{-4}$ & $2{\times}10^{-4}$ & $(0.5,\,0.999)$ & BCE (real=0.9) & $0.999$ \\
\midrule
\multirow{3}{*}{StyleGAN2-lite}
& MNIST        & 100 & 128 & 128 & 64 & $2{\times}10^{-4}$ & $2{\times}10^{-4}$ & $(0.0,\,0.99)$  & R1 ($\gamma{=}10$) & $0.999$ \\
& FashionMNIST & 120 & 128 & 128 & 64 & $2{\times}10^{-4}$ & $2{\times}10^{-4}$ & $(0.0,\,0.99)$  & R1 ($\gamma{=}10$) & $0.999$ \\
& CIFAR-10     & 200 & 128 & 128 & 64 & $2{\times}10^{-4}$ & $2{\times}10^{-4}$ & $(0.0,\,0.99)$  & R1 ($\gamma{=}10$) & $0.999$ \\
\bottomrule
\end{tabular}
\end{adjustbox}
\end{table}

\begin{table}[!htbp]
\centering
\footnotesize
\caption{\textbf{DCGAN} and \textbf{StyleGAN2-lite} architectures. 
$C{=}1$ (MNIST/Fashion-MNIST), $C{=}3$ (CIFAR-10).}
\label{tab:gans-arch}
\begin{adjustbox}{max width=\textwidth}
\begin{tabularx}{\textwidth}{@{}p{0.28\textwidth}X@{}}
\toprule
\textbf{DCGAN (cond.)} &
\textbf{G:} $[z|\,\mathrm{emb}(y)] \to$ FC $\to$ seed ($7\times 7$ if $28\times 28$, else $8\times 8$), width $g_\mathrm{ch}{=}128$; two ConvT $4\times 4$, $s{=}2$ (BN+ReLU) to $g_\mathrm{ch}/4$; final $3\times 3$ conv $\to C$; $\tanh$. 
\newline
\textbf{D:} concat one-hot (10 ch) on input; Conv $4\times 4$, $s{=}2$: $C{+}10 \to d_\mathrm{ch}/2 \to d_\mathrm{ch}$ (BN+LReLU 0.2); final conv $k{=}7/8 \to$ logit.\\
\midrule
\textbf{StyleGAN2-lite (cond., $32\times 32$)} &
\textbf{G:} MLP on $[z|\,\mathrm{emb}(y)] \to w$ (256-d, $4\times$ LReLU); learned $4\times 4\times g_\mathrm{ch}$ constant; StyledConv (mod+demod, per-layer noise) with upsampling $4\to 8\to 16\to 32$; skip-sum \texttt{ToRGB} at 4/8/16/32; $\tanh$. 
\newline
\textbf{D:} \texttt{FromRGB} $\to$ residual downs $32\to 16\to 8\to 4$ (avg-pool); final convs; projection: scalar $+ \langle \mathrm{proj}(h), \mathrm{emb}(y)\rangle$ (256-d).\\
\bottomrule
\end{tabularx}
\end{adjustbox}
\end{table}

\paragraph{Conditional diffusion models (cDDPM).}
We train both UNet- and Transformer2D-based cDDPMs with classifier-free guidance~\citep{ho2022classifier} and evaluate using
fast samplers at test time. Training and model configurations appear in
Tables~\ref{tab:cddpm-train}--\ref{tab:unetddpm-model}.

\begin{table}[!htbp]
\centering
\scriptsize
\caption{Optimization and diffusion schedule for \textbf{cDDPM} (all datasets). Data/splits follow Table~\ref{tab:img-data-shared}.}
\label{tab:cddpm-train}
\begin{adjustbox}{max width=\linewidth}
\begin{tabular}{@{}lccccccccc@{}}
\toprule
Datasets & Epochs & Batch & Optim & LR & Timesteps & Pred.\ type & EMA & Clip & Seed \\
\midrule
MNIST / FashionMNIST / CIFAR-10 & 50 & 128 & AdamW & $1{\times}10^{-4}$ & 1000 & v-pred & 0.999 & 1.0 & 42 \\
\bottomrule
\end{tabular}
\end{adjustbox}
\end{table}

\begin{table}[!htbp]
\centering
\scriptsize
\caption{UNet config, conditioning, and sampling for \textbf{cDDPM} (only $C/H/W$ vary by dataset).}
\label{tab:cddpm-model}
\begin{adjustbox}{max width=\linewidth}
\begin{tabular}{@{}lcccccccc@{}}
\toprule
UNet blocks & Stage ch. & Heads & Cond dim & $p_{\text{uncond}}$ & Sampler & Steps & CFG & Preview \\
\midrule
$3\downarrow / 3\uparrow$, 2 layers (cross-attn) & (128, 256, 256) & 4 & 128 & 0.10 & DPM-Solver++ & 50 & 2.0 & $10$/class \\
\bottomrule
\end{tabular}
\end{adjustbox}
\end{table}

\begin{table}[!htbp]
\centering
\scriptsize
\caption{Training hyperparameters for \textbf{Transformer2D cDDPM}. Optimizer: AdamW $(\beta_1{=}0.9,\beta_2{=}0.999)$; loss: MSE on $v$ (v-prediction).}
\label{tab:t2d-train}
\begin{adjustbox}{max width=\linewidth}
\begin{tabular}{@{}lccccccccc@{}}
\toprule
Dataset & Epochs & Batch & LR & Weight decay & Timesteps & Pred.\ type & EMA decay & Sample steps & Seed \\
\midrule
MNIST        & 150 & 256 & $1{\times}10^{-4}$ & 0.0 & 1000 & v-pred & 0.9999 & 16 & 42 \\
FashionMNIST & 160 & 256 & $1{\times}10^{-4}$ & 0.0 & 1000 & v-pred & 0.9999 & 20 & 42 \\
CIFAR-10     & 400 & 256 & $1{\times}10^{-4}$ & 0.0 & 1000 & v-pred & 0.9999 & 20 & 42 \\
\bottomrule
\end{tabular}
\end{adjustbox}
\end{table}

\begin{table}[!htbp]
\centering
\scriptsize
\caption{Model and conditioning for \textbf{Transformer2D cDDPM}. Training uses a DDPM scheduler with linear $\beta$ ($10^{-4}\!\to\!2\!\times\!10^{-2}$). Sampling uses DPM-Solver++ with classifier-free guidance~\citep{lu2022dpmsolverpp}.}
\label{tab:t2d-model}
\begin{adjustbox}{max width=\linewidth}
\begin{tabular}{@{}lccccccccc@{}}
\toprule
Dataset & Patch & Layers & Heads & Head dim & Cond dim & $p_{\text{uncond}}$ & CFG scale & In ch & Size \\
\midrule
MNIST        & 2 & 8  & 4 & 64 & 128 & 0.20 & 2.0 & 1 & 28 \\
FashionMNIST & 2 & 10 & 4 & 64 & 128 & 0.20 & 2.0 & 1 & 28 \\
CIFAR-10     & 4 & 12 & 6 & 64 & 128 & 0.20 & 2.0 & 3 & 32 \\
\bottomrule
\end{tabular}
\end{adjustbox}
\end{table}

\begin{table}[!htbp]
\centering
\scriptsize
\caption{Model and sampling configuration for \textbf{UNet cDDPM}. Training scheduler: DDPM (squared-cosine $\beta$, v-pred); sampler: Euler Ancestral with CFG.}
\label{tab:unetddpm-model}
\begin{adjustbox}{max width=\linewidth}
\begin{tabular}{@{}lcccccccc@{}}
\toprule
Dataset & Blocks (ch.) & Layers/block & In/Out ch. & Size & Class slots & Cond. type & Pred.\ type & Sampler \\
\midrule
MNIST / FashionMNIST & (128, 256, 256) & 2 & $1\to1$ & 28 & $K{+}1$ (0=null) & class\_embed (timestep) & v-pred & EulerA \\
CIFAR-10        & (128, 256, 256)\;{\small or}\;(128, 256, 512)\textsuperscript{$\dagger$} & 2 & $3\to3$ & 32 & $K{+}1$ (0=null) & class\_embed (timestep) & v-pred & EulerA \\
\bottomrule
\end{tabular}
\end{adjustbox}

\vspace{2pt}\footnotesize
$\dagger$ \texttt{--wide} switches CIFAR-10 to $(128,256,512)$; \texttt{--channels} accepts a custom comma list (e.g., \texttt{128,256,256}).\\
During training, classifier-free guidance drops labels to the null class with probability $p_{\text{uncond}}$. At sampling, the doubled batch uses labels $0$ and $c+1$ for the unconditional and conditional predictions. With guidance scale $s$, the combined prediction is $v_{\mathrm{uncond}}+s(v_{\mathrm{cond}}-v_{\mathrm{uncond}})$.
\end{table}

\FloatBarrier

\section{Sensitivity, Cost, and Scaling}
\label{appendix:sensitivity-cost-scaling}

Section~\ref{sec:synthetic-quantile} describes the expected responses to the
controlled perturbations. Appendix~\ref{app:standard-metric-comparison}
reports the correlations from the matched comparison, and
Appendix~\ref{appendix:Experiments} gives the settings for the separate
one-factor sensitivity analysis. The sensitivity experiments vary dataset size, classifier, scoring rule, and
class balance; they do not vary the neighborhood size $k$ in Similarity
or the number of train/test splits.

The computational cost of \method{} includes repeated classifier fitting,
evaluation resampling, and nearest-neighbor computations. In the synthetic
CPU experiments with $n=10{,}000$, 10 resampling iterations, and a tree-based
classifier, a full configuration takes approximately $2.03$--$2.04$ minutes.
This timing characterizes the specified configuration; a comparative runtime
benchmark is not reported.

The image experiments use MNIST, Fashion-MNIST, and CIFAR-10 for
class-conditional evaluation. Evaluating computational scaling on larger
image datasets, such as CelebA and ImageNet, remains a direction for future work.

\section{Proxy-Supervised Evaluation}
\label{appendix:proxy-supervision}

\method uses a supervised task to probe a property of generated samples. Labels may already encode a property of interest, such as molecular activity, image category, or text topic. When the generated and real samples come without labels, the evaluator can define a proxy task that encodes a distinction to examine.

An outlier detector can create a binary task by labeling samples as inliers or outliers. This asks whether generated data preserves the structure needed to separate typical from atypical examples. Clustering can create a multiclass task by assigning samples to clusters and using the cluster indices as proxy labels. This lets the evaluator probe whether generated samples preserve coarse regions of the real data. Reporting how the proxy labels were assigned makes clear which distinction the scores examine.

Quality measures transfer on the proxy task, Utility measures additional predictive value on that task, and Similarity measures mixing within the proxy classes. Indistinguishability uses the real-versus-generated discriminator. The diagnostic meaning of the proxy-dependent scores follows from the distinction that the proxy labels encode.

Table~\ref{tab:proxy-labels} gives a small illustration. We generate $10{,}000$ samples with 40 features and assign proxy inlier/outlier labels using an Isolation Forest. We then simulate collapse by progressively replacing samples with duplicates. In this proxy task, Quality and Utility remain saturated: the duplicated samples still support the inlier/outlier decision. By contrast, Indistinguishability and Exchangeability decrease monotonically as collapse increases. The example illustrates how proxy supervision extends the protocol to
datasets without natural labels. The resulting scores reflect the distinctions
encoded by the chosen proxy task.

\begin{table}[!htbp]
\centering
\small
\setlength{\tabcolsep}{6pt}
\caption{Proxy-supervised collapse experiment. Samples receive inlier/outlier proxy labels from an Isolation Forest before mode collapse is simulated by increasing duplication.}
\label{tab:proxy-labels}
\begin{tabular}{ccccc}
\toprule
\textbf{Collapse} & \textbf{Q} & \textbf{U} & \textbf{I} & \textbf{E} \\
\midrule
0.0 & 1.00 & 1.00 & 0.24 & 0.12 \\
0.2 & 1.00 & 1.00 & 0.20 & 0.10 \\
0.4 & 1.00 & 1.00 & 0.14 & 0.07 \\
0.6 & 1.00 & 1.00 & 0.09 & 0.05 \\
0.8 & 1.00 & 1.00 & 0.05 & 0.03 \\
1.0 & 1.00 & 1.00 & 0.00 & 0.00 \\
\bottomrule
\end{tabular}
\end{table}

\FloatBarrier

\section{Molecular Generator Ranks by Dataset}
\begin{figure}[!htbp]
  \centering
  \includegraphics[width=1.0\linewidth]{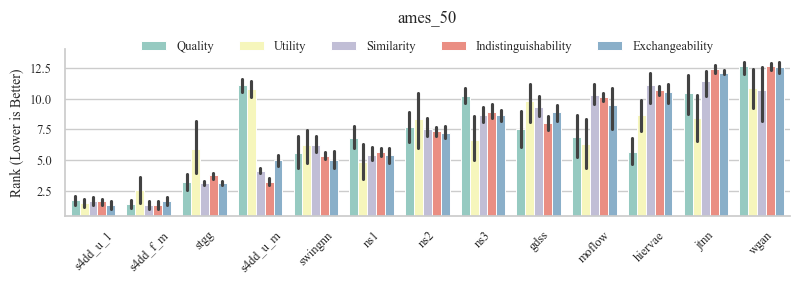}
  \caption{Average generator rank on \textbf{AMES} across vectorizers (lower is better).}
  \label{fig:rank_ames}
\end{figure}

\begin{figure}[!htbp]
  \centering
  \includegraphics[width=1.0\linewidth]{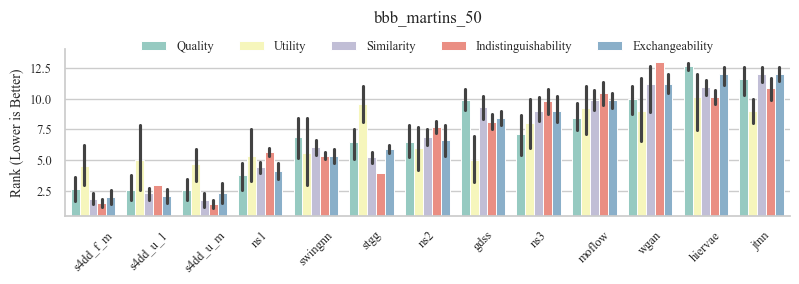}
  \caption{Average generator rank on \textbf{BBB Martins} across vectorizers (lower is better).}
  \label{fig:rank_bbb}
\end{figure}

\begin{figure}[!htbp]
  \centering
  \includegraphics[width=1.0\linewidth]{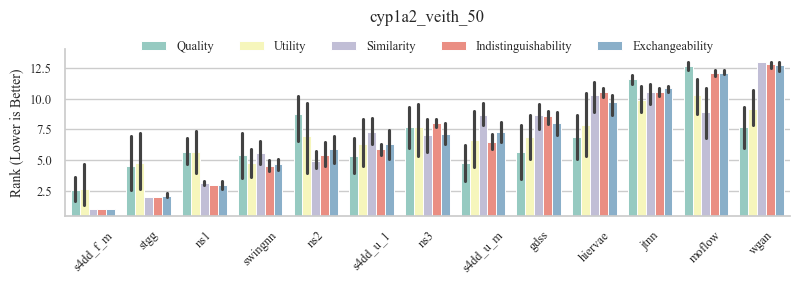}
  \caption{Average generator rank on \textbf{CYP1A2} across vectorizers (lower is better).}
  \label{fig:rank_cypa12}
\end{figure}

\begin{figure}[!htbp]
  \centering
  \includegraphics[width=1.0\linewidth]{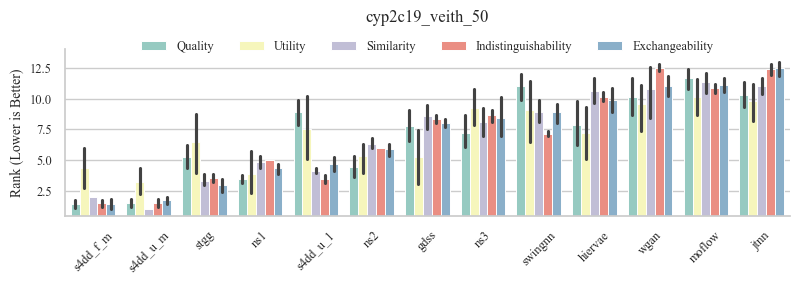}
  \caption{Average generator rank on \textbf{CYP2C19} across vectorizers (lower is better).}
  \label{fig:rank_cyp2c19}
\end{figure}

\begin{figure}[!htbp]
  \centering
  \includegraphics[width=1.0\linewidth]{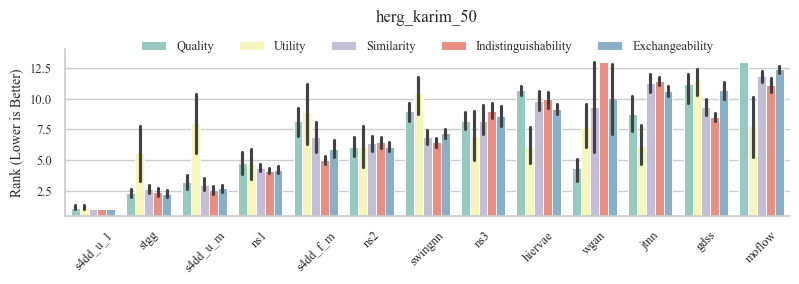}
  \caption{Average generator rank on \textbf{hERG Karim} across vectorizers (lower is better).}
  \label{fig:rank_herg_karim}
\end{figure}

Figures~\ref{fig:rank_ames}--\ref{fig:rank_herg_karim} show the molecular generator ranks for each dataset and vectorizer. Lower ranks indicate better scores.

% =========================================================
% Image-Domain Experiments
% =========================================================
\FloatBarrier

\section{Image-Domain Experiments: Datasets, Models, and Full Results}
\label{app:image-experiments}
% (Shared preprocessing and split policy retained.)

\subsection{Datasets, Preprocessing, and Splits}
\label{app:image-data}
See Table~\ref{tab:img-data-shared} for the shared image preprocessing and split policy.

\begin{table}[!htbp]
\centering
\scriptsize
\caption{Shared preprocessing and canonical splits used for all models and experiments.}
\label{tab:img-data-shared}
\begin{adjustbox}{max width=\linewidth}
\begin{tabular}{@{}lcccc@{}}
\toprule
Dataset & Shape $(C\times H\times W)$ & Pixel scale & Default train aug. & Split policy \\
\midrule
MNIST & $1\times28\times28$ & $[-1,1]$ from $[0,1]$ & none & stratified 50/50 \texttt{train1}/\texttt{train2} (seed 42); test untouched \\
FashionMNIST & $1\times28\times28$ & $[-1,1]$ & none & same \\
CIFAR-10 & $3\times32\times32$ & $[-1,1]$ & none$^\dagger$ & same \\
\bottomrule
\end{tabular}
\end{adjustbox}

\vspace{2pt}\footnotesize $^\dagger$ Some models enable flip/crop at training time; exports match dataset size.
\end{table}

% (Model/training tables retained; minor wording normalized.)

\FloatBarrier

\subsection{Image Data Full Metric Results Tables}
\label{app:image-results}
See Tables~\ref{tab:vals-ranks-cifar10}--\ref{tab:ke-mnist-ranked} for the full image-domain \method and conventional metric results.
For the three GUIDE tables, $\Emin$ is computed from the displayed component
scores as $\min\{(\Q+\U)/2,(\Ind+\Simi)/2\}$. These plug-in summaries may
differ from the median of the per-resample composite scores. The corresponding
$\Emin$ ranks are computed in descending order, assigning average ranks to ties
in the displayed-score calculations.

% --- CIFAR-10 ---
\begin{table}[!htbp]\centering\scriptsize
\caption{Metric values (left) and ranks (right) on \textbf{CIFAR-10}. Lower rank is better.}
\label{tab:vals-ranks-cifar10}
\begin{adjustbox}{max width=\linewidth}
\begin{tabular}{l rr rr rr rr rr}
\toprule
\multirow{2}{*}{\textbf{Generator}} & \multicolumn{2}{c}{\textbf{Q}} & \multicolumn{2}{c}{\textbf{U}} & \multicolumn{2}{c}{\textbf{S}} & \multicolumn{2}{c}{\textbf{I}} & \multicolumn{2}{c}{$\Emin$} \\
 & \textit{val} & \textit{rank} & \textit{val} & \textit{rank} & \textit{val} & \textit{rank} & \textit{val} & \textit{rank} & \textit{val} & \textit{rank} \\
\midrule
DCGAN & 0.670 & 9.0 & 0.000 & 6.5 & 0.450 & 4.0 & 0.450 & 1.0 & 0.335 & 4.0 \\
DDPM & 0.940 & 2.5 & 0.250 & 3.0 & 0.550 & 1.0 & 0.430 & 2.0 & 0.490 & 1.0 \\
StyleGAN2 & 0.880 & 4.0 & 0.000 & 6.5 & 0.190 & 9.0 & 0.080 & 6.0 & 0.135 & 7.0 \\
Transformer2D & 0.950 & 1.0 & 0.290 & 1.0 & 0.510 & 3.0 & 0.400 & 4.0 & 0.455 & 3.0 \\
UNet\_CFG & 0.940 & 2.5 & 0.250 & 2.0 & 0.530 & 2.0 & 0.410 & 3.0 & 0.470 & 2.0 \\
VAE\_Conv & 0.760 & 6.0 & 0.000 & 6.5 & 0.230 & 7.0 & 0.010 & 7.5 & 0.120 & 8.0 \\
VAE\_MLP & 0.740 & 7.0 & 0.000 & 6.5 & 0.210 & 8.0 & 0.010 & 7.5 & 0.110 & 9.0 \\
VAE\_Res & 0.770 & 5.0 & 0.000 & 6.5 & 0.400 & 5.0 & 0.000 & 9.0 & 0.200 & 6.0 \\
WGAN-GP & 0.720 & 8.0 & 0.000 & 6.5 & 0.280 & 6.0 & 0.140 & 5.0 & 0.210 & 5.0 \\
\bottomrule
\end{tabular}
\end{adjustbox}
\end{table}

% --- Fashion-MNIST ---
\begin{table}[!htbp]\centering\scriptsize
\caption{Metric values (left) and ranks (right) on \textbf{Fashion-MNIST}. Lower rank is better.}
\label{tab:vals-ranks-fashionmnist}
\begin{adjustbox}{max width=\linewidth}
\begin{tabular}{l rr rr rr rr rr}
\toprule
\multirow{2}{*}{\textbf{Generator}} & \multicolumn{2}{c}{\textbf{Q}} & \multicolumn{2}{c}{\textbf{U}} & \multicolumn{2}{c}{\textbf{S}} & \multicolumn{2}{c}{\textbf{I}} & \multicolumn{2}{c}{$\Emin$} \\
 & \textit{val} & \textit{rank} & \textit{val} & \textit{rank} & \textit{val} & \textit{rank} & \textit{val} & \textit{rank} & \textit{val} & \textit{rank} \\
\midrule
DCGAN & 0.930 & 5.5 & 0.020 & 1.0 & 0.300 & 6.0 & 0.100 & 3.0 & 0.200 & 5.5 \\
DDPM & 0.940 & 2.5 & 0.000 & 5.5 & 0.500 & 1.5 & 0.170 & 1.0 & 0.335 & 1.0 \\
StyleGAN2 & 0.940 & 2.5 & 0.000 & 5.5 & 0.220 & 7.0 & 0.000 & 7.0 & 0.110 & 7.0 \\
Transformer2D & 0.810 & 8.0 & 0.000 & 5.5 & 0.060 & 8.0 & 0.000 & 7.0 & 0.030 & 8.5 \\
UNet\_CFG & 0.940 & 2.5 & 0.000 & 5.5 & 0.500 & 1.5 & 0.140 & 2.0 & 0.320 & 2.0 \\
VAE\_Conv & 0.920 & 7.0 & 0.000 & 5.5 & 0.400 & 5.0 & 0.000 & 7.0 & 0.200 & 5.5 \\
VAE\_MLP & 0.940 & 2.5 & 0.000 & 5.5 & 0.430 & 3.5 & 0.000 & 7.0 & 0.215 & 3.5 \\
VAE\_Res & 0.930 & 5.5 & 0.000 & 5.5 & 0.430 & 3.5 & 0.000 & 7.0 & 0.215 & 3.5 \\
WGAN-GP & 0.790 & 9.0 & 0.000 & 5.5 & 0.050 & 9.0 & 0.010 & 4.0 & 0.030 & 8.5 \\
\bottomrule
\end{tabular}
\end{adjustbox}
\end{table}
% ===== MNIST =====
\begin{table}[!htbp]\centering\scriptsize
\caption{Metric values (left) and ranks (right) on \textbf{MNIST}. Lower rank is better.}
\label{tab:vals-ranks-mnist}
\begin{adjustbox}{max width=\linewidth}
\begin{tabular}{l rr rr rr rr rr}
\toprule
\multirow{2}{*}{\textbf{Generator}} & \multicolumn{2}{c}{\textbf{Q}} & \multicolumn{2}{c}{\textbf{U}} & \multicolumn{2}{c}{\textbf{S}} & \multicolumn{2}{c}{\textbf{I}} & \multicolumn{2}{c}{$\Emin$} \\
 & \textit{val} & \textit{rank} & \textit{val} & \textit{rank} & \textit{val} & \textit{rank} & \textit{val} & \textit{rank} & \textit{val} & \textit{rank} \\
\midrule
dcgan         & 0.560 & 9.0 & 0.000 & 7.0 & 0.010 & 9.0 & 0.000 & 6.0 & 0.005 & 9.0 \\
ddpm          & 0.970 & 3.0 & 0.000 & 7.0 & 0.570 & 2.0 & 0.130 & 1.5 & 0.350 & 2.0 \\
stylegan2     & 0.920 & 7.0 & 0.090 & 1.0 & 0.090 & 7.0 & 0.060 & 6.0 & 0.075 & 7.0 \\
transformer2d & 0.820 & 8.0 & 0.000 & 7.0 & 0.070 & 8.0 & 0.000 & 6.0 & 0.035 & 8.0 \\
unet\_cfg     & 0.970 & 3.0 & 0.000 & 7.0 & 0.580 & 1.0 & 0.130 & 1.5 & 0.355 & 1.0 \\
vae\_conv     & 0.960 & 6.0 & 0.000 & 7.0 & 0.480 & 5.0 & 0.000 & 6.0 & 0.240 & 5.0 \\
vae\_mlp      & 0.970 & 3.0 & 0.000 & 7.0 & 0.490 & 4.0 & 0.000 & 6.0 & 0.245 & 4.0 \\
vae\_res      & 0.970 & 3.0 & 0.000 & 4.0 & 0.520 & 3.0 & 0.000 & 6.0 & 0.260 & 3.0 \\
wgangp        & 0.970 & 3.0 & 0.090 & 2.0 & 0.400 & 6.0 & 0.000 & 6.0 & 0.200 & 6.0 \\
\bottomrule
\end{tabular}
\end{adjustbox}
\end{table}

% --- CIFAR-10 (FID/KID/PR/IS) ---
\begin{table}[!htbp]
\centering
\small
\caption{CIFAR-10 metrics with per-metric ranks ($r{=}1$ best). $n_{\text{ref}}{=}25{,}000$, $n_{\text{gen}}{=}25{,}000$.}
\label{tab:cifar10-ranked}
\begin{adjustbox}{max width=\linewidth}
\begin{tabular}{lrrrrrrrrrrrr}
\toprule
\textbf{Model} & \textbf{FID} $\downarrow$ & \textbf{r} & \textbf{KID} $\downarrow$ & \textbf{r} & \textbf{P} & \textbf{r} & \textbf{R} & \textbf{r} & \textbf{F1\_pr} & \textbf{r} & \textbf{IS} & \textbf{r} \\
\midrule
DCGAN         & 21.62 & 5 & 0.0171 & 5 & 0.165 & 6 & 0.078 & 3 & 0.106 & 3 & 2.041 & 1 \\
DDPM          &  6.67 & 1 & 0.0030 & 1 & 0.224 & 3 & 0.122 & 1 & 0.158 & 1 & 2.017 & 2 \\
StyleGAN2     & 19.26 & 3 & 0.0129 & 3 & 0.220 & 4 & 0.002 & 6 & 0.005 & 6 & 1.958 & 3 \\
UNet\_CFG     &  9.62 & 2 & 0.0069 & 2 & 0.288 & 1 & 0.093 & 2 & 0.141 & 2 & 1.944 & 4 \\
Transformer2D & 20.13 & 4 & 0.0150 & 4 & 0.286 & 2 & 0.065 & 4 & 0.106 & 3 & 1.763 & 5 \\
VAE\_Conv     & 182.67 & 7 & 0.1905 & 7 & 0.110 & 7 & 0.000 & 7 & 0.000 & 7 & 1.293 & 7 \\
VAE\_MLP      & 206.49 & 8 & 0.2182 & 8 & 0.094 & 8 & 0.000 & 7 & 0.000 & 7 & 1.280 & 8 \\
VAE\_Res      & 374.51 & 9 & 0.4439 & 9 & 0.002 & 9 & 0.000 & 7 & 0.000 & 7 & 1.006 & 9 \\
WGAN-GP        &  42.62 & 6 & 0.0342 & 6 & 0.207 & 5 & 0.017 & 5 & 0.032 & 5 & 1.652 & 6 \\
\bottomrule
\end{tabular}
\end{adjustbox}
\end{table}

% --- Fashion-MNIST (FID/KID/PR/IS) ---
\begin{table}[!htbp]
\centering
\small
\caption{Fashion-MNIST metrics with per-metric ranks ($r{=}1$ best). $n_{\text{ref}}{=}30{,}000$, $n_{\text{gen}}{=}30{,}000$.}
\label{tab:fashionmnist-ranked}
\begin{adjustbox}{max width=\linewidth}
\begin{tabular}{lrrrrrrrrrrrr}
\toprule
\textbf{Model} & \textbf{FID} $\downarrow$ & \textbf{r} & \textbf{KID} $\downarrow$ & \textbf{r} & \textbf{P} & \textbf{r} & \textbf{R} & \textbf{r} & \textbf{F1\_pr} & \textbf{r} & \textbf{IS} & \textbf{r} \\
\midrule
DCGAN         & 0.19 & 1 & 0.0106 & 4 & 0.215 & 7 & 0.174 & 3 & 0.192 & 3 & 2.364 & 1 \\
DDPM          & 0.71 & 2 & 0.0018 & 1 & 0.369 & 2 & 0.246 & 1 & 0.295 & 1 & 1.994 & 3 \\
StyleGAN2     & 1.45 & 4 & 0.0077 & 3 & 0.318 & 3 & 0.001 & 8 & 0.002 & 8 & 1.869 & 5 \\
UNet\_CFG     & 1.10 & 3 & 0.0042 & 2 & 0.463 & 1 & 0.188 & 2 & 0.268 & 2 & 1.905 & 4 \\
Transformer2D & 5.45 & 9 & 0.0587 & 9 & 0.090 & 8 & 0.001 & 8 & 0.002 & 8 & 1.570 & 9 \\
VAE\_Conv     & 3.82 & 6 & 0.0231 & 6 & 0.230 & 6 & 0.009 & 4 & 0.017 & 4 & 1.692 & 6 \\
VAE\_MLP      & 4.59 & 8 & 0.0216 & 5 & 0.252 & 5 & 0.007 & 5 & 0.014 & 5 & 1.677 & 7 \\
VAE\_Res      & 3.99 & 7 & 0.0257 & 7 & 0.257 & 4 & 0.006 & 6 & 0.011 & 6 & 1.673 & 8 \\
WGAN-GP        & 2.19 & 5 & 0.0423 & 8 & 0.057 & 9 & 0.005 & 7 & 0.009 & 7 & 2.013 & 2 \\
\bottomrule
\end{tabular}
\end{adjustbox}
\end{table}

% --- MNIST (FID/KID/PR/IS) ---
\begin{table}[!htbp]
\centering
\small
\caption{MNIST metrics with per-metric ranks ($r{=}1$ best). $n_{\text{ref}}{=}30{,}000$, $n_{\text{gen}}{=}30{,}000$.}
\label{tab:mnist-ranked}
\begin{adjustbox}{max width=\linewidth}
\begin{tabular}{lrrrrrrrrrrrr}
\toprule
\textbf{Model} & \textbf{FID} $\downarrow$ & \textbf{r} & \textbf{KID} $\downarrow$ & \textbf{r} & \textbf{P} & \textbf{r} & \textbf{R} & \textbf{r} & \textbf{F1\_pr} & \textbf{r} & \textbf{IS} & \textbf{r} \\
\midrule
DCGAN         & 3.89 & 9 & 0.0321 & 8 & 0.307 & 3 & 0.003 & 7 & 0.007 & 7 & 1.276 & 9 \\
DDPM          & 0.25 & 4 & 0.0016 & 1 & 0.488 & 2 & 0.319 & 1 & 0.386 & 1 & 1.634 & 7 \\
StyleGAN2     & 2.27 & 8 & 0.0144 & 6 & 0.291 & 4 & 0.000 & 9 & 0.000 & 9 & 1.478 & 8 \\
UNet\_CFG     & 0.29 & 5 & 0.0017 & 2 & 0.531 & 1 & 0.299 & 2 & 0.383 & 2 & 1.636 & 6 \\
Transformer2D & 0.83 & 7 & 0.1345 & 9 & 0.027 & 9 & 0.003 & 7 & 0.005 & 8 & 1.647 & 5 \\
VAE\_Conv     & 0.24 & 3 & 0.0114 & 3 & 0.223 & 5 & 0.072 & 5 & 0.109 & 5 & 1.784 & 3 \\
VAE\_MLP      & 0.23 & 2 & 0.0125 & 4 & 0.206 & 8 & 0.082 & 4 & 0.117 & 4 & 1.816 & 2 \\
VAE\_Res      & 0.40 & 6 & 0.0141 & 5 & 0.212 & 6 & 0.052 & 6 & 0.084 & 6 & 1.716 & 4 \\
WGAN-GP        & 0.14 & 1 & 0.0163 & 7 & 0.210 & 7 & 0.207 & 3 & 0.209 & 3 & 1.865 & 1 \\
\bottomrule
\end{tabular}
\end{adjustbox}
\end{table}

% --- Kernel-entropy diversity proxies (CIFAR-10) ---
\begin{table}[!htbp]
\centering
\small
\caption{Kernel-entropy metrics on CIFAR-10 with per-metric ranks ($r{=}1$ best). Higher is better for RKE\_gen, FKEA-VENDI\_gen, FKEA-RKE2\_gen; lower is better for RRKE.}
\label{tab:ke-cifar10-ranked}
\begin{adjustbox}{max width=\linewidth}
\begin{tabular}{lrrrrrrrrrrr}
\toprule
\textbf{Model} & $\boldsymbol{\sigma}$ & \textbf{RKE\_gen} & \textbf{r} & $\boldsymbol{\hat m}$ & \textbf{RRKE} & \textbf{r} & \textbf{FKEA-VENDI\_gen} & \textbf{r} & \textbf{FKEA-RKE2\_gen} & \textbf{r} \\
\midrule
DCGAN         & 15.18 & 0.852 & 5 & 2 & 0.179 & 5 & 13.1 & 3 & 2.3 & 3 \\
DDPM          & 15.72 & 0.942 & 1 & 3 & 0.120 & 1 & 15.7 & 1 & 2.6 & 1 \\
StyleGAN2     & 15.21 & 0.858 & 3 & 2 & 0.167 & 4 & 12.6 & 4 & 2.3 & 3 \\
UNet\_CFG     & 15.54 & 0.901 & 2 & 2 & 0.129 & 2 & 14.0 & 2 & 2.4 & 2 \\
Transformer2D & 15.28 & 0.855 & 4 & 2 & 0.165 & 3 & 12.4 & 5 & 2.4 & 2 \\
VAE\_Conv     & 15.34 & 0.335 & 7 & 1 & 0.762 & 7 &  2.8 & 7 & 1.4 & 5 \\
VAE\_MLP      & 15.54 & 0.305 & 8 & 1 & 0.829 & 8 &  2.5 & 8 & 1.4 & 5 \\
VAE\_Res      & 17.04 & 0.014 & 9 & 1 & 1.228 & 9 &  1.1 & 9 & 1.0 & 6 \\
WGAN-GP        & 14.82 & 0.745 & 6 & 2 & 0.249 & 6 &  9.5 & 6 & 2.1 & 4 \\
\bottomrule
\end{tabular}
\end{adjustbox}
\end{table}

% --- Kernel-entropy diversity proxies (Fashion-MNIST) ---
\begin{table}[!htbp]
\centering
\small
\caption{Kernel-entropy metrics on Fashion-MNIST with per-metric ranks ($r{=}1$ best).}
\label{tab:ke-fashionmnist-ranked}
\begin{adjustbox}{max width=\linewidth}
\begin{tabular}{lrrrrrrrrrrr}
\toprule
\textbf{Model} & $\boldsymbol{\sigma}$ & \textbf{RKE\_gen} & \textbf{r} & $\boldsymbol{\hat m}$ & \textbf{RRKE} & \textbf{r} & \textbf{FKEA-VENDI\_gen} & \textbf{r} & \textbf{FKEA-RKE2\_gen} & \textbf{r} \\
\midrule
DCGAN         & 16.63 & 0.953 & 1 & 3 & 0.120 & 3 & 14.7 & 1 & 2.6 & 1 \\
DDPM          & 16.56 & 0.945 & 2 & 3 & 0.083 & 1 & 13.1 & 2 & 2.6 & 1 \\
StyleGAN2     & 16.39 & 0.905 & 4 & 2 & 0.133 & 4 & 11.1 & 4 & 2.5 & 2 \\
UNet\_CFG     & 16.46 & 0.923 & 3 & 3 & 0.092 & 2 & 11.9 & 3 & 2.5 & 2 \\
Transformer2D & 16.62 & 0.848 & 7 & 2 & 0.357 & 9 &  8.7 & 9 & 2.3 & 4 \\
VAE\_Conv     & 16.19 & 0.852 & 6 & 2 & 0.205 & 6 &  9.3 & 6 & 2.3 & 4 \\
VAE\_MLP      & 16.26 & 0.852 & 6 & 2 & 0.199 & 5 &  9.0 & 8 & 2.3 & 4 \\
VAE\_Res      & 16.12 & 0.837 & 8 & 2 & 0.211 & 7 &  9.1 & 7 & 2.3 & 4 \\
WGAN-GP        & 16.54 & 0.879 & 5 & 2 & 0.263 & 8 & 10.6 & 5 & 2.4 & 3 \\
\bottomrule
\end{tabular}
\end{adjustbox}
\end{table}

% --- Kernel-entropy diversity proxies (MNIST) ---
\begin{table}[!htbp]
\centering
\small
\caption{Kernel-entropy metrics on MNIST with per-metric ranks ($r{=}1$ best).}
\label{tab:ke-mnist-ranked}
\begin{adjustbox}{max width=\linewidth}
\begin{tabular}{lrrrrrrrrrrr}
\toprule
\textbf{Model} & $\boldsymbol{\sigma}$ & \textbf{RKE\_gen} & \textbf{r} & $\boldsymbol{\hat m}$ & \textbf{RRKE} & \textbf{r} & \textbf{FKEA-VENDI\_gen} & \textbf{r} & \textbf{FKEA-RKE2\_gen} & \textbf{r} \\
\midrule
DCGAN         & 13.58 & 0.869 & 8 & 2 & 0.331 & 8 &  8.1 & 8 & 2.3 & 5 \\
DDPM          & 13.48 & 0.971 & 3 & 3 & 0.070 & 1 & 13.5 & 2 & 2.7 & 1 \\
StyleGAN2     & 13.21 & 0.882 & 7 & 2 & 0.235 & 7 &  8.7 & 7 & 2.4 & 4 \\
UNet\_CFG     & 13.40 & 0.972 & 2 & 3 & 0.071 & 2 & 13.0 & 3 & 2.6 & 2 \\
Transformer2D & 15.28 & 0.850 & 9 & 2 & 0.540 & 9 & 10.7 & 6 & 2.3 & 5 \\
VAE\_Conv     & 13.46 & 0.937 & 5 & 3 & 0.146 & 4 & 12.5 & 4 & 2.5 & 3 \\
VAE\_MLP      & 13.67 & 0.953 & 4 & 3 & 0.150 & 5 & 13.0 & 3 & 2.6 & 2 \\
VAE\_Res      & 13.38 & 0.921 & 6 & 3 & 0.161 & 6 & 11.9 & 5 & 2.5 & 3 \\
WGAN-GP        & 13.93 & 0.994 & 1 & 3 & 0.145 & 3 & 15.0 & 1 & 2.7 & 1 \\
\bottomrule
\end{tabular}
\end{adjustbox}
\end{table}

\FloatBarrier
\paragraph{Relationship between \method and conventional metrics.}
Tables~\ref{tab:vals-ranks-cifar10}--\ref{tab:vals-ranks-mnist} report \method metrics. Tables~\ref{tab:cifar10-ranked}--\ref{tab:ke-mnist-ranked} report FID, KID, precision, recall, Inception Score, and kernel-entropy measures under the selected image representations. On CIFAR-10, DDPM and UNet--CFG rank first and second on both FID and KID and score highly on several GUIDE probes. On MNIST, both rank among the top two on KID and Similarity, with Utility reported as $0.000$. StyleGAN2 combines competitive CIFAR-10 FID and KID with low recall, Utility, and Similarity. WGAN--GP has Utility reported as $0.000$ on CIFAR-10 and $0.090$ on MNIST. The profiles show how the comparisons vary by criterion and dataset.

\subsection{Reproducibility Notes}
Evaluation uses equal $n_{\text{ref}}$ and $n_{\text{gen}}$ per dataset (see captions).

% =============================
% End of Appendix
% =============================

\clearpage

\section{Examples from the Molecular and Image Generators}
\label{appendix:example}
\begin{figure}[H]
    \centering
    \includegraphics[width=\textwidth]{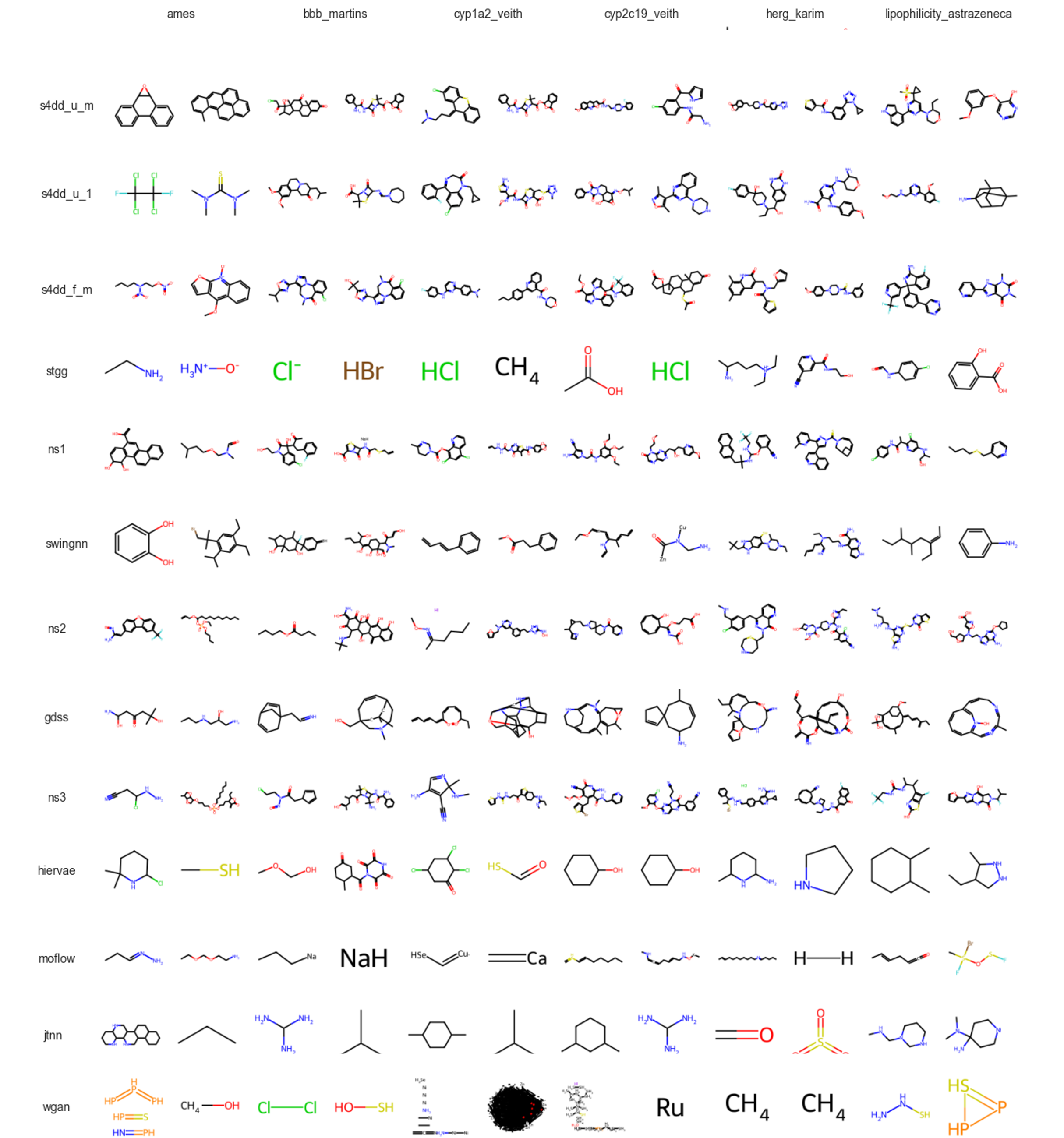}
    \caption{Examples of molecules generated by different models across the six datasets. Each row is a generator; each dataset contributes a positive and a negative sample.}
    \label{fig:generated_molecules}
\end{figure}

\begin{figure}[H]
  \centering
  \includegraphics[width=0.65\linewidth]{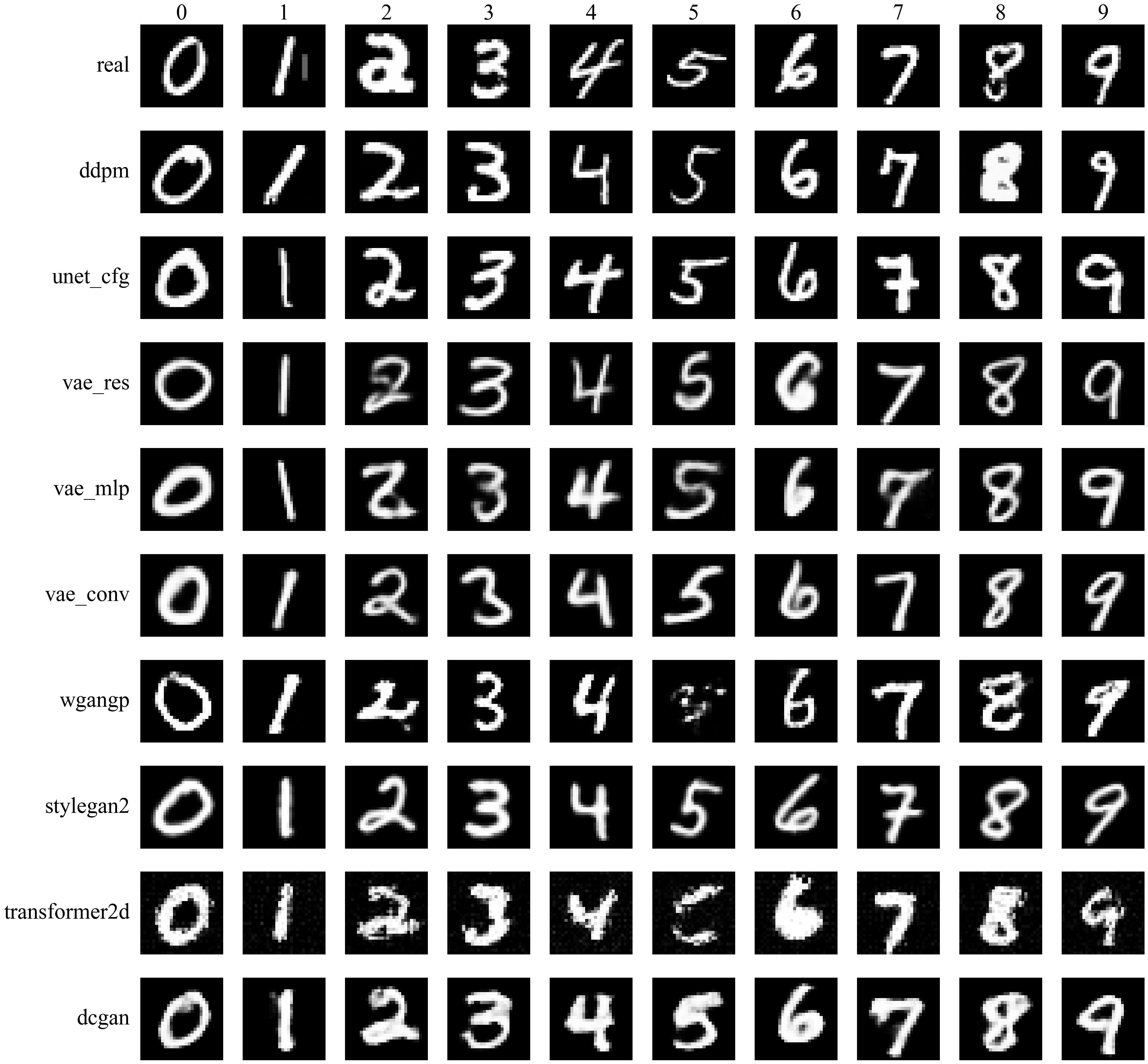}
  \caption{Samples generated by each model on MNIST. Rows: generators; columns: class-conditional samples (uncurated).}
  \label{fig:mnist_generated_panel}
\end{figure}

\begin{figure}[H]
  \centering
  \includegraphics[width=0.65\linewidth]{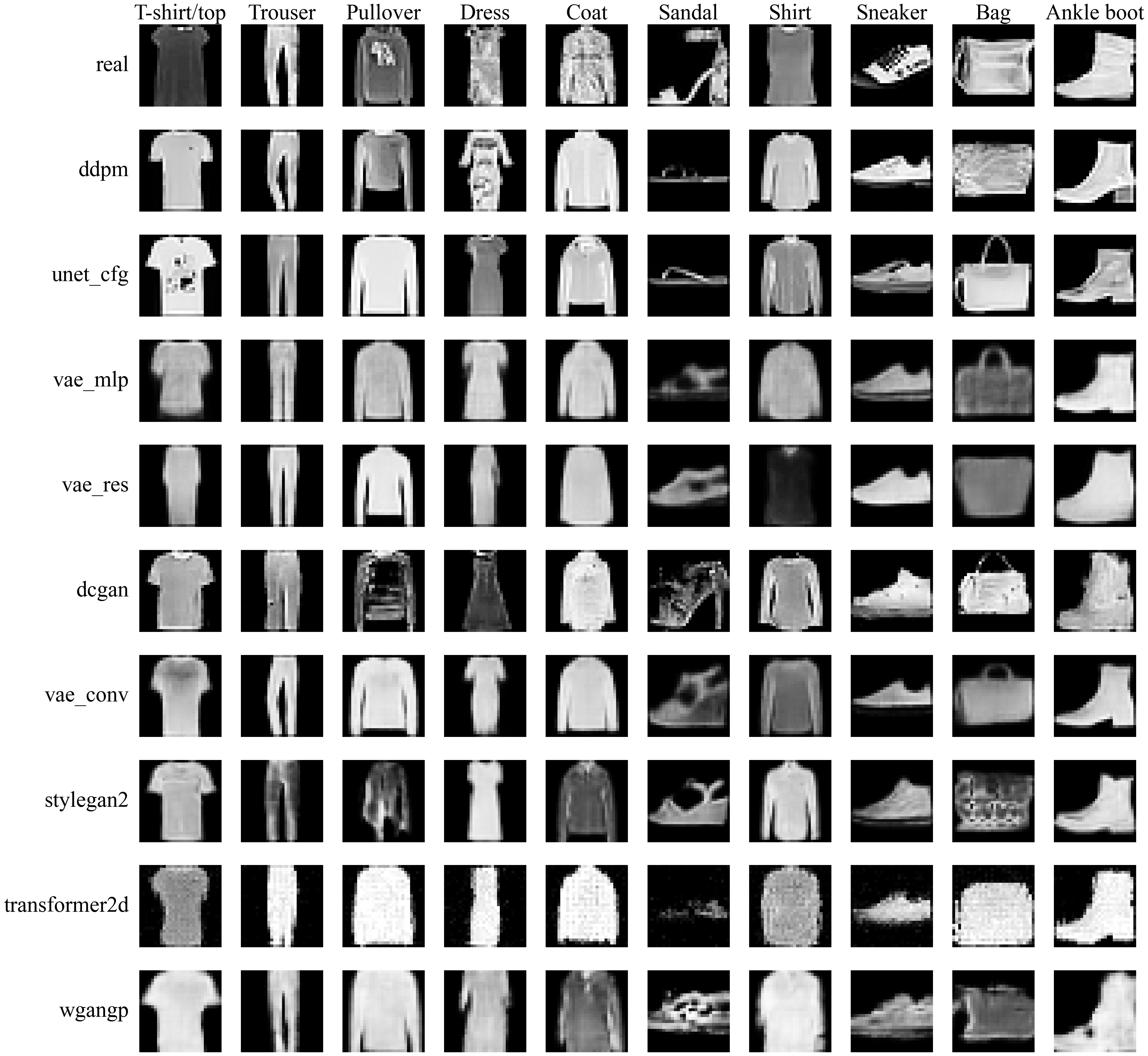}
  \caption{Samples generated by each model on Fashion-MNIST. Rows: generators; columns: class-conditional samples (uncurated).}
  \label{fig:fashionmnist_generated_panel}
\end{figure}

\begin{figure}[H]
  \centering
  \includegraphics[width=0.94\linewidth]{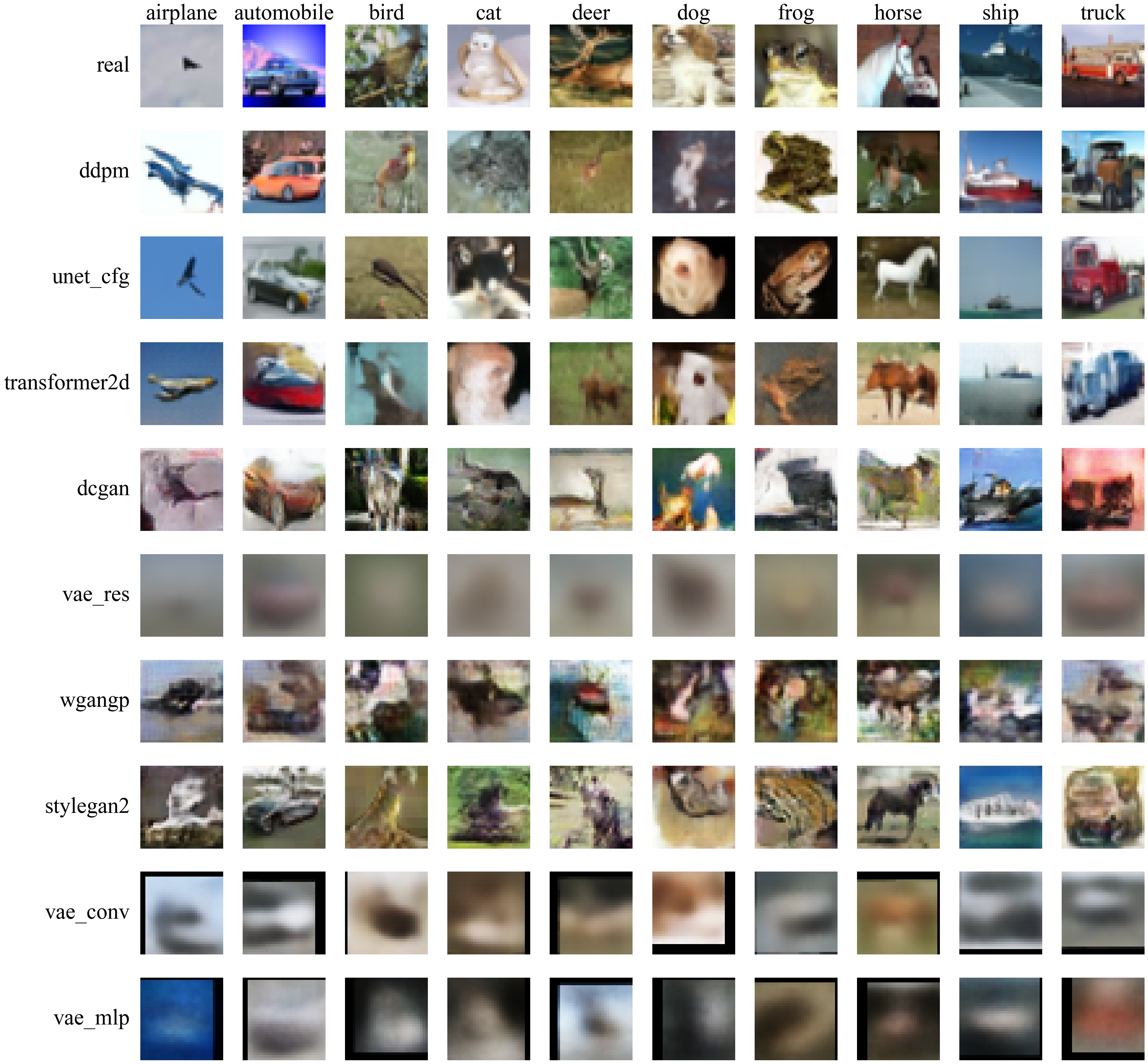}
  \caption{Samples generated by each model on CIFAR-10. Rows: generators; columns: per-class conditional samples (uncurated).}
  \label{fig:cifar10_generated_panel}
\end{figure}

See Figures~\ref{fig:generated_molecules}--\ref{fig:cifar10_generated_panel} for uncurated samples from the molecular and image generators.

t